%% file: arxiv.tex
\documentclass[10pt,twocolumn,letterpaper]{article}

\usepackage[pagenumbers]{cvpr} 


\AtEndPreamble

\definecolor{cvprblue}{rgb}{0.21,0.49,0.74}
\usepackage[pagebackref,breaklinks,colorlinks,allcolors=cvprblue]{hyperref}
\usepackage{bbm}
\usepackage{multirow} 
\usepackage{float}
\newcommand{\model}{VN$\circlearrowleft$Bert}

\title{NOLO: Navigate Only Look Once}

\author{
  Bohan Zhou \\
  School of Computer Science \\
  Peking University \\
  {\tt\small zhoubh@stu.pku.edu.cn}
  \and
  Zhongbin Zhang \\
  Department of Automation \\
  Tsinghua University \\
  {\tt\small zhangzb23@mails.tsinghua.edu.cn} 
  \and
  Jiangxing Wang \\
  School of Computer Science \\
  Peking University \\
  {\tt\small jiangxiw@stu.pku.edu.cn} 
  \and
  Zongqing Lu\thanks{Corresponding author}\\
  School of Computer Science \\
  Peking University \\
  BAAI \\
  {\tt\small zongqing.lu@pku.edu.cn}
}

\begin{document}
\maketitle
\input{sec/0_abstract}    
\input{sec/1_intro}
\input{sec/2_relatedwork}
\input{sec/3_method}
\input{sec/4_experiment}
\input{sec/5_conclusion}

{
    \small
    \bibliographystyle{ieeenat_fullname}
    \bibliography{reference}
}

\input{sec/X_suppl}

\end{document}

%% file: sec/0_abstract.tex
\begin{abstract}
The in-context learning ability of Transformer models has brought new possibilities to visual navigation. In this paper, we focus on the \textbf{video navigation} setting, where an in-context navigation policy needs to be learned purely from videos in an offline manner, without access to the actual environment. For this setting, we propose \textbf{N}avigate \textbf{O}nly \textbf{L}ook \textbf{O}nce (NOLO), a method for learning a navigation policy that possesses the in-context ability and adapts to new scenes by taking corresponding context videos as input without finetuning or re-training. To enable learning from videos, we first propose a pseudo action labeling procedure using optical flow to recover the action label from egocentric videos. Then, offline reinforcement learning is applied to learn the navigation policy. Through extensive experiments on different scenes both in simulation and the real world, we show that our algorithm outperforms baselines by a large margin, which demonstrates the in-context learning ability of the learned policy. For videos and more information, visit our \href{https://sites.google.com/view/nol0}{project page}.
\end{abstract}

%% file: sec/1_intro.tex
\section{Introduction}
\label{sec:intro}
\input{figs/head}

Visual navigation has long become a research focus due to its wide application in different fields, including mobile robots~\cite{intro1}, autonomous vehicles~\cite{intro2}, and virtual assistants~\cite{intro3}. 
Existing approaches to visual navigation can be broadly divided into two main categories: modular and end-to-end methods. Modular methods decompose the navigation problem into specific tasks, such as simultaneous localization and mapping, object segmentation, keypoint matching, and depth estimation~\cite{ANS, NTS, GOAT, OVRL, MOD-INN}. End-to-end methods~\cite{VLN-Bert, VINT, VLN1}, on the other hand, bypass such modularization, directly learning the mappings from raw sensory observations (e.g., GPS and Compass~\cite{NORW, pointgoal}, RGB-D cameras~\cite{NRNS}, and IMU~\cite{imu-nav}) to action outputs. While these approaches have yielded promising results, they still face significant challenges, particularly in terms of generalizability. Modular methods can be brittle when transferred to unfamiliar environments. End-to-end approaches mainly focus on online learning~\cite{NTS,xu2023vision,chen2024scale}, which requires continuous, interactive engagement with simulated or real environments. Thus, they may require additional interactions or exploration to adapt to novel scenes, complicating their deployment in real-world navigation settings.

To address these limitations, it is helpful to consider how humans intuitively navigate. We can often find our way through unfamiliar spaces after watching a simple traversal video. Our navigation abilities do not depend on precise positional data or distance measurement. Instead, we rely primarily on visual recognition, understanding, and memory.
Building on this insight, we introduce a new visual navigation setting called \textbf{Video Navigation}, which mirrors how humans rely on visual observations to navigate. 
In video navigation, a mobile agent is provided with a context video of an unfamiliar environment and learns to reach any target seen in the video. For instance, a sweeping robot equipped with video navigation capabilities could immediately navigate and clean any area within a house after viewing a short video of its layout. We believe video navigation has immense potential for endowing mobile agents with human-like navigation capabilities. Although encouraging, achieving robust video navigation poses considerable challenges: 
\begin{itemize}[itemsep=0.1em,parsep=0em,topsep=0em,partopsep=0em]
    \item Limited Observations. Learning to navigate from egocentric traversal videos is challenging due to a lack of information, including actual actions, spatial cues, scene structure, and visual context.
    \item Absence of Explicit Intent. Egocentric videos lack clear navigational intentions, making conventional methods like inverse reinforcement learning unsuitable, as these approaches typically require expert trajectory data.
    \item Minimal Sensory Input. Camera intrinsics, robot poses, maps, odometers, or depth inputs are not necessary for video navigation, which distinguishes our work from others. A majority of existing methods prefer to build explicit spatial maps for planning with heterogeneous sensors. In this work, we aim to enable human-like navigation using only egocentric RGB observations.
\end{itemize}

Fundamentally, the objective of video navigation is to train a meta-policy that enables a robot to adapt to a novel scene--essentially, to navigate anywhere using cues from a context video. Recent research demonstrates that in-context learning (ICL~\cite{InContextSurvey}) excels at solving tasks using a few contextual demonstrations without requiring additional training~\cite{GPT3} and has proven valuable across fields from LLM-based agents~\cite{LLMagents} to meta-learning~\cite{AD,DPT,AT}. Inspired by the success of ICL, we propose \textbf{N}avigate \textbf{O}nly \textbf{L}ook \textbf{O}nce \textbf{(NOLO)}, which learns a transferable in-context navigation policy conditioned on a context video, enabling adaptation to new scenes without fine-tuning or further re-training. 

To achieve this, we collect a dataset of egocentric traversal videos, each covering a unique scene. We derive pseudo-actions from these context videos using optical flow, generating full frame-action trajectories, and then employ an offline reinforcement learning approach to train the navigation policy. This resource-efficient offline training pipeline eliminates the need for time-consuming online interactions, enabling large-scale training on simulated data and facilitating the transfer of navigation skills to unfamiliar environments. Furthermore, NOLO’s in-context adaptability makes it particularly suitable for real-world deployment, where frequent policy adjustments are often impractical.





\vspace{2mm}
\noindent Our key contributions can be summarized as follows:
\begin{itemize}[itemsep=0.1em,parsep=0em,topsep=0em,partopsep=0em]
    \item We present a novel and practical setting \textbf{Video Navigation}, which necessitates learning an in-context navigation policy in an offline manner purely from videos such that the learned policy can adapt to different scenes by taking corresponding videos as context.
    \item We propose \textbf{NOLO} to solve the video navigation problem. NOLO seamlessly incorporates optical flow into offline reinforcement learning via pseudo-action labeling. To encourage the temporal coherence of representation, we further propose a temporal coherence loss for temporally aligned context visual representation.
    \item Empirical evaluations on \textbf{RoboTHOR} and \textbf{Habitat} benchmarks demonstrate the advantages of NOLO over baselines in video navigation. Notably, navigation is achieved by only viewing a single 30-second video clip per scene. We further deploy NOLO on the Unitree Go2 robot and experiments verify that NOLO is also effective and friendly to real-world deployment.
\end{itemize}

%% file: figs/head.tex
\begin{figure*}[h]
\centering
\includegraphics[width=.76\textwidth]{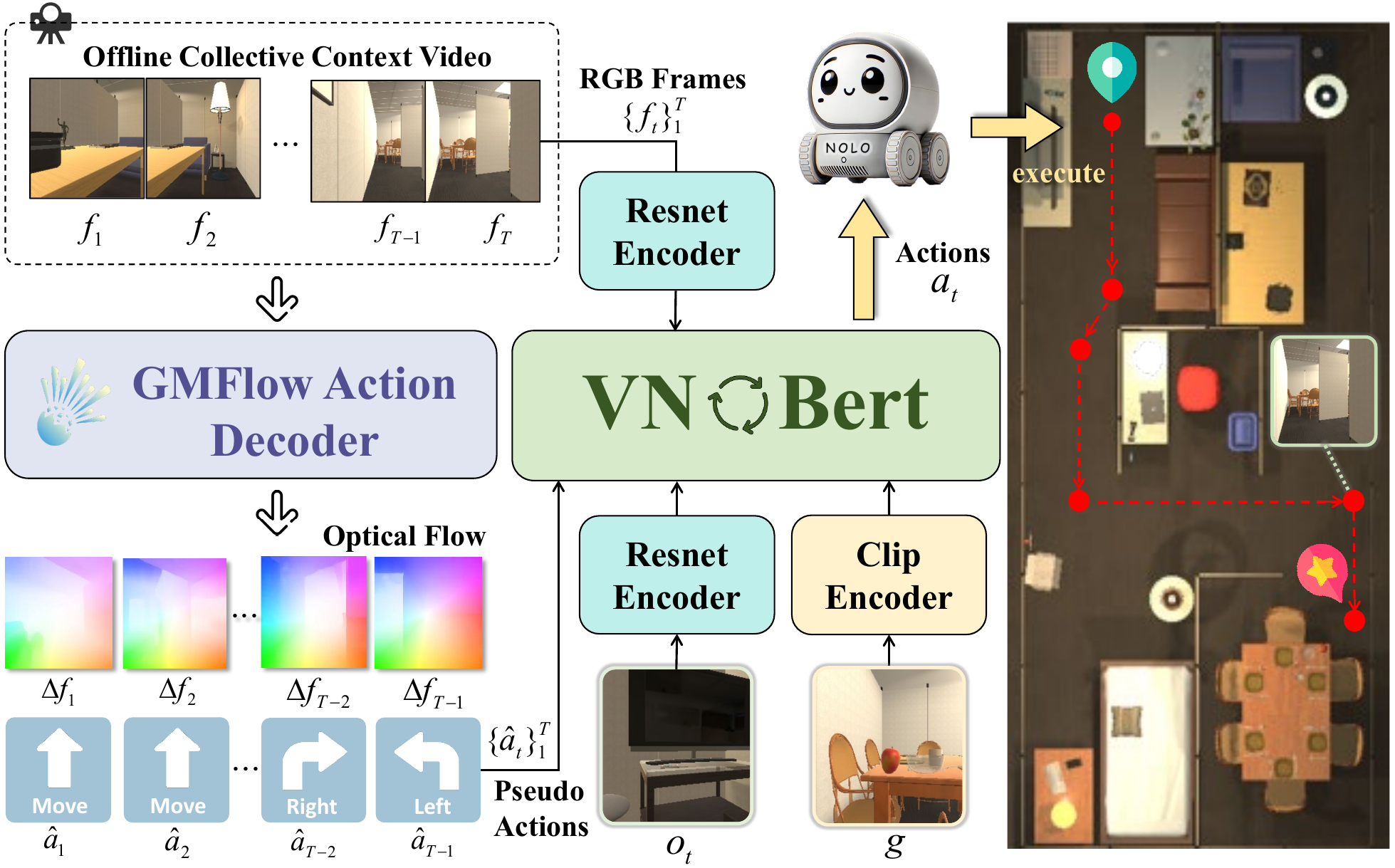}
\setlength{\belowcaptionskip}{-12pt}
\setlength{\textfloatsep}{10pt} 

\caption{The framework of NOLO. 
An offline collected egocentric video is taken by a pretrained GMFlow action decoder to label pseudo action sequence $\{\hat{a}_t\}_1^{T-1}$. An in-context video navigation policy $\pi_{\theta}(\cdot|g,o_t,\{f_t\}_1^T,\{\hat{a}_t\}_1^{T-1})$, modeled by \model, is learned to take all context frames $\{f_t\}_1^T$, labeled actions $\{\hat{a}_t\}_1^{T-1}$, a current observation $o_t$ and a goal image $g$ to generate discrete actions to navigate to the desired target in a novel scene.
}
\label{fig:head} 
\end{figure*}


%% file: sec/2_relatedwork.tex
\section{Related Work}
\label{sec:related}

\textbf{Visual Navigation.}
The visual navigation problem can be categorized into indoor navigation~\cite{scale-VLN,Airbert,NaVid,georgakis2019simultaneous,li2020learning,mayo2021visual,kwon2021visual,VINT,Zson,kim2023topological,li2024memonav} and outdoor navigation~\cite{VLN-VIDEO,LM-Nav,wasserman2023last}.
In this paper, we only focus on the indoor navigation problem. Current research mainly focuses on finding the location of specific positions~\cite{pointgoal,EmbCLIP}, rooms~\cite{Memory-Augmented}, or object instances~\cite{ISIGN,MOD-INN,EmbCLIP,IEVE}. The goals of navigation can be described in coordinates~\cite{pointgoal}, natural language~\cite{Poni,OVRL,Ovrl-v2,VLN1,VLN2}, goal image of target position~\cite{Memory-Augmented,OVRL,Ovrl-v2}, or goal image of object instances~\cite{ISIGN,MOD-INN,EmbCLIP,IEVE,AVDC,Zson,kwon2021visual}. Our video navigation of finding objects occurred in the context video falls into the last group.
Visual navigation methods can also be categorized by whether exploration is required. Most existing works aim for generalization across new goals or environments, heavily relying on learning algorithms that require extensive online interactions, often through reinforcement learning~\cite{imgNavBirth, FGPrompt, ZER}, paired with real-time visual feedback~\cite{mayo2021visual}. This approach enables the agent to adapt dynamically but requires significant interaction within the environment. Some studies attempt to improve data efficiency but still rely on random policy exploration or assume access to rich datasets, including robot poses~\cite{GOAT} or depth images~\cite{NRNS,georgakis2019simultaneous,kim2023topological,MOD-INN}. In contrast, NOLO learns entirely offline. It leverages video data for navigation, freeing the agent from real-world exploration and allowing effective policy learning without performing actual actions, marking a novel approach in video-based visual navigation.


\noindent\textbf{Learning from Videos.}
There are four main lines of research study learning from videos. (1) Learn distinctive representations for intrinsic rewards. These approaches mainly include predicting the future~\cite{AFP-RL,VIPER,STG}, learning temporal abstractions~\cite{STG,TCN,TDC,ELE}, contrastive learning~\cite{CURL,UL}, adversarial learning~\cite{AMP,VGAIfO,VGAIfO-SO,VIL} or other transition-learning based methods~\cite{AILOT,TDRP,HILPs}.
(2) Recover value function along with representations~\cite{NCE,VIP,ICVF,V-PTR}.
(3) Pretrain action-free model in the first stage and then boost the model with additional action information~\cite{FICC,APV,ContextWM}.
(4) Discover latent actions~\cite{ILPO,DePO,LAPO,Giene} and then align to the real actions online in the environment~\cite{ILPO,DePO,LAPO}, by human annotation~\cite{Giene} or from offline datasets~\cite{LAPO}.
Unlike previous work, we focus on the visual navigation problem and by utilizing its specificity, propose an offline method that directly learns the actual navigation policy, not the latent one, from videos. 
Few attempts are devoted to decoding real actions directly from videos. A recent work AVDC~\cite{AVDC} attempts to principally use similar rule-based methods to directly infer actions for robot manipulation. However, it is an online algorithm with additional depth image inputs. We also compare NOLO with AVDC~\cite{AVDC} in \cref{sec:ablation}.

\noindent\textbf{In-Context Learning.}
In-context learning (ICL), first demonstrated in large language models~\cite{GPT3, llama, palm}, enables models to perform new tasks with a few examples as context. This capability has extended to vision-language models~\cite{flamingo, Frozen, GPT4}, AI agents~\cite{hugginggpt, metagpt}, and models across heterogeneous modalities~\cite{graphICL, PointCloudICL}. Our work uniquely applies ICL to video navigation. To our knowledge, NOLO is the first attempt to learn an adaptable policy guided by context videos for diverse scenes.

%% file: sec/3_method.tex
\section{Learning to Navigate from Videos}
\subsection{The Problem Formulation of Video Navigation}
\label{sec:Formulation}
In video navigation, we offline recover a navigation policy $\pi_{\theta}\left(a|g,o,\mathcal{V} \right)$ to find objects occurred in the context video, which takes a goal image $g$, current observation $o$, and context video $\mathcal{V}$ as input and outputs discrete action $a$, considering high-level discrete actions are commonly used in competitions like the Habitat Navigation Challenge~\cite{Habitat-web} and modern navigation tasks~\cite{VLN-VIDEO}. To train such a policy, we assume a video dataset $\mathcal{D}$ contains $N$ videos collected (e.g., by some simple rule-based policy or humans) in $N$ different scenes, $\mathcal{D} =\left\{ \mathcal{V}^1,\mathcal{V}^2,...,\mathcal{V}^N \right\}$. Each video $\mathcal{V}^{i}$ contains $T$ frames $f^i_t$, $\mathcal{V}^i =\left\{ f^i_1,f^i_2,...,f^i_T \right\}$. During inference, we deploy the learned policy in a new scene given a context video $\mathcal{V^\textrm{new}} \notin \mathcal{D}$ and a goal frame $g\in\mathcal{G}^\textrm{new}$ describing an object occurred in $\mathcal{V^\textrm{new}}$. The policy is expected to adapt to this new scene via in-context learning.

\subsection{Pseudo Action Labeling}
\label{sec:Pseudo-Action-Labeling}

In video navigation, capturing environmental dynamics is crucial for effective adaptation to new scenes. Building on prior work~\cite{IUPE,ILPO,DePO,LAPO,Giene} that leverages inverse dynamics models (IDMs) to extract latent actions from adjacent frames, we make a further step by converting latent actions into actual discrete actions within video navigation. Specifically, we propose pseudo action labeling, as illustrated in ~\cref{fig:opticalflow}.
we employ GMFlow\cite{Gmflow} as an implicit IDM $F_\xi$ to estimate discrete pixel displacements between consecutive frames and predict real navigation actions from flow maps with rule-based post-processing. 
For each context video $\mathcal{V}$, we label a sequence of pseudo-actions $\boldsymbol{\hat{a}}$ to get a trajectory $\mathcal{T}=\{f_t,\hat{a}_t\}^{T}_{t=1}$. 
See Appendix~\ref{appendix:Pseudo} for more details.



\input{figs/opticalflow1}

\subsection{In-Context Policy Modeling}
After labeling pseudo action for all videos, we can build a frame-action dataset to offline train an in-context navigation policy $\pi_{\theta}$. Inspired by VLN-Bert~\cite{VLN-Bert}, we propose a bidirectional recurrent Transformer, \textbf{\model}, to model the in-context navigation policy $\pi_\theta$, 
which can be formalized in \cref{equ:VNBert}:
\begin{equation}
\label{equ:VNBert}
    h_{t+1}, \pi_{\theta}\left({a}_t|g,o_t,\mathcal{T}\right) = \text{VN}\circlearrowleft\text{Bert}(g,o_t,\mathcal{T}),
\end{equation}
where $h$ is specially designed as a recurrent hidden state to capture informative attention history. 

The structure of \model is depicted in \cref{fig:structure}. For inference, all RGB images, including context frames and the current observation are encoded by a learnable ResNet encoder $E_\zeta$ pretrained from Places365~\cite{Places365} $e^s_t=E_\zeta(f_t)$. Actions are encoded by a learnable action embedding layer $e^a_t=E_\alpha(\hat{a}_t)$ and the goal frame is encoded by a fixed pretrained CLIP~\cite{CLIP} $e^g=\text{CLIP}(g)$.
During initialization, a zero-padded hidden state $h_\text{init}$ is additionally inserted at the back of the context embedding sequence for recurrence. The entire token embedding sequence $\boldsymbol{e}_{\text{init}}=[h_\text{init},e^s_{\text{init},1},e^a_{\text{init},1}\dots,e^a_{\text{init},T-1},e^s_{\text{init},T}]$ is encoded by a multi-layer bidirectional self-attention module (SA) as $\boldsymbol{e}_0=[\boldsymbol{e}^c,h_0]=\text{SA}(\boldsymbol{e}_{\text{init}})$. The output feature of the hidden state $h_0$ serves as the initial compressed representation of the context trajectory. 
After initialization, the context embedding sequence $\boldsymbol{e}^c$ keeps unchanged. At each timestep $t$, the current observation $o_t$ and the goal image $g$ are encoded into $e^s_{t}$ and $e^g$, respectively. The concatenated embedding sequence $\boldsymbol{e}_t=[h_t,\boldsymbol{e}^c,e^s_t,e^g]$ is then fed into the cross-attention module (CA) to obtain the updated embedding $\boldsymbol{e}_{t+1}=\text{CA}(\boldsymbol{e}_t)$ which is further aggregated and projected to action distribution $\pi_{\theta}\left({a}_t|g,o_t,\mathcal{T} \right)$ and the next hidden state $h_{t+1}$. The updated hidden state $h_{t+1}$ is subsequently processed by a multilayer perceptron (MLP) $P_Q$ to regress Q-values $Q_\omega\left(g,o_t,a_t,\mathcal{T} \right)$ 
and by another MLP $P_D$ to predict the binary terminal signal $\delta_\upsilon\left(d_t|g,o_t,\mathcal{T} \right)$, where $d_t$ is used for predicting the termination of the current task. 

\input{figs/structure1}

To train {\model}, we sample a context video $\mathcal{V}^i$ from dataset $\mathcal{D}$ and then construct a context trajectory $\mathcal{T}^i=\{f^i_t,\hat{a}^i_t\}^{T}_{t=1}$ as described in Section \ref{sec:Pseudo-Action-Labeling}. Then,
we randomly sample $t\in[0,T)$ for a current observation $o_t$ and pseudo action $\hat{a}_t$, and a goal image $g\in\mathcal{G}^i$.
We set $d_t = \mathbbm{I}(o_t=g)$ for the termination prediction, where $\mathbbm{I}$ is indicator function. We forward the model and calculate discrete action prediction loss and termination prediction loss in \cref{eq:a},
\begin{equation}
\label{eq:a}
\begin{split}
\mathcal{L}_{a}=-\mathbb{E}_{g,o_t, \hat{a}_t\sim \mathcal{T}^i}
\log \pi_{\theta}\left(\hat{a}_t|g,o_t,\mathcal{T}^i \right)
\\
\mathcal{L}_{d}=-\mathbb{E}_{g,o_t\sim \mathcal{T}^i}
\log \delta_\upsilon\left(d_t|g,o_t,\mathcal{T}^i \right).
\end{split}
\end{equation}

\subsection{Batch-Constrained Q-Learning}
\label{sec:bcq}
Since the policy used to record the video may be highly suboptimal for video navigation, we favor offline reinforcement learning over imitation learning for policy learning. Specifically, we adopt BCQ~\cite{BCQ}, an offline reinforcement learning method that emphasizes constraining the action selection to those within the distribution of the observed frame-action pairs. Q-function $Q_\omega$ is learned from Bellman update in \cref{eq:BCQ1}, where $r(g, o_t, \hat{a}_t) = \mathbbm{I}(o_t=g)$ is a binary reward indicating the success of a task,
\begin{equation}
\label{eq:BCQ1}
\begin{split}
\mathcal{L}_q=\mathop{\mathbb{E}}_{o_t, o_{t+1}, g, \hat{a}_t \sim \mathcal{T}^i}\left[ Q_{\omega}\left( g,o_t,\hat{a}_t,\mathcal{T}^i \right) - y_t \right]^2, 
\\
y_t=r(g, o_t, \hat{a}_t) + \gamma \max_{\Tilde{a}_{t+1} \in A^\beta}Q_{\omega}\left( g,o_{t+1},\Tilde{a}_{t+1},\mathcal{T}^i \right).
\end{split}
\end{equation}

Given the action space is discrete, we define a set $A^\beta$ in \cref{eq:BCQ2} to select $\Tilde{a}_{t+1}$, where $\beta$ represents the threshold for action selection. Action choices are constrained based on the action distribution within the context trajectory dataset. The learned Q-function, $Q_\omega$, thus can wisely guides the agent toward the target goal $g$.
\begin{equation}
\label{eq:BCQ2}
    A^\beta := \Big\{\Tilde{a}_{t+1}\Big|\frac{\pi _{\theta}\left( \cdot|g,o_{t+1},\mathcal{T}^i \right)}{\max \pi _{\theta}\left( \cdot|g,o_{t+1},\mathcal{T}^i \right)} >\beta \Big\}.
\end{equation}

\subsection{Temporal Coherence}
\label{sec:Temporal Preference}
Learning temporally aligned representation of videos is crucial for understanding the causality of the context video. Inspired by RLHF~\cite{Preference}, we learn a "fuzzy" measure of progress from their natural temporal ordering, where the higher preference score is assigned to the later frames in the video. To achieve this, we learn a temporal indicator $\hat{u}_\phi(e)$, which consists of 1D self-attention and spectral normalization modules to map a visual context embedding $e^s_t$ to a utility score. $\hat{u}_\phi$ is optimized according to \cref{eq:temporal}: 
\begin{equation}
\label{eq:temporal}
\min_{\zeta,\phi}\mathcal{L}_{t}=\underset{f_{t_-}, f_{t_+} \sim \mathcal{V}}{\mathbb{E}}
-\log \sigma\left[\hat{u}_\phi\left(E_\zeta(f_{t_+}) \right)-\hat{u}_\phi\left(E_\zeta(f_{t_-})\right) \right],
\end{equation}
where $t_+ > t_-$ and $f_{t_+}, f_{t_-}$ stands for the later and earlier frame in the video. This self-supervised paradigm naturally drives the temporal indicator $\hat{u}_\phi(s)$ to maximize likelihood to ensure the temporal order preference $f_{t_+}\succ f_{t_-}$, leading to temporally aligned context visual representation, which is beneficial to better understand the temporal relation in a novel scene, as elaborated in later \cref{sec:ablation}.

%% file: figs/opticalflow1.tex
\begin{figure*}
\centering
\includegraphics[width=0.8\linewidth]{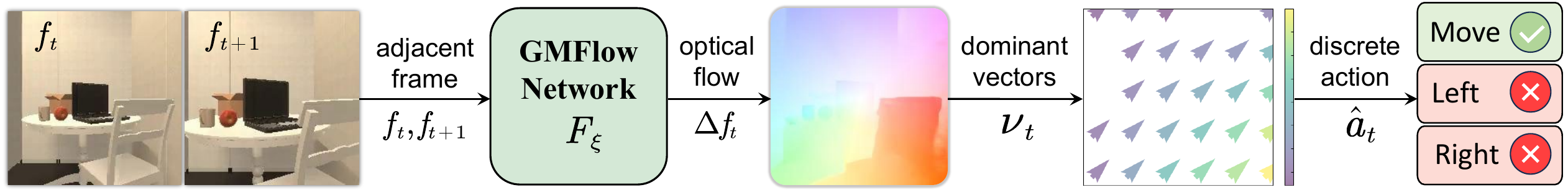}
\setlength{\textfloatsep}{10pt} 
\setlength{\belowcaptionskip}{-10pt}
  
\caption{Two adjacent frames are taken by a pretrained optical flow model to get a flow map. Some representative dominant vectors are filtered for action selection.}
\label{fig:opticalflow} 
\end{figure*}

%% file: figs/structure1.tex
\begin{figure*}[t]
\centering
\includegraphics[width=\textwidth]{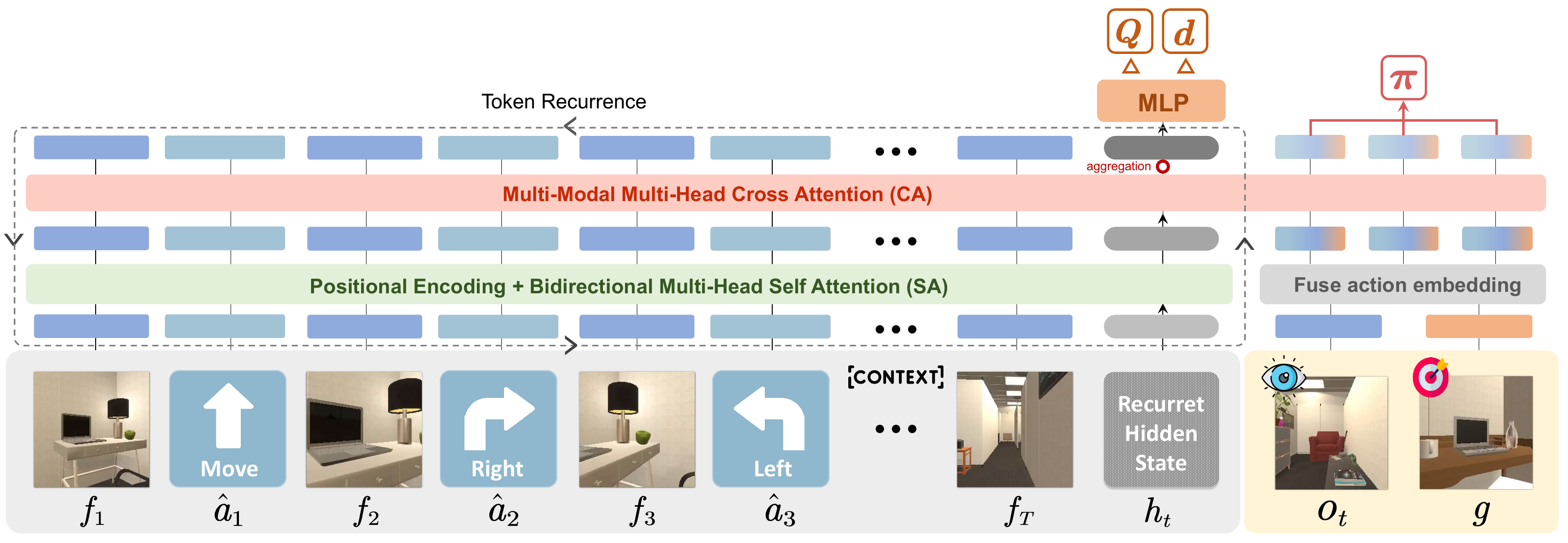}
\setlength{\abovecaptionskip}{-2pt} 
\setlength{\belowcaptionskip}{-8pt}  
\caption{Structure of \model. At the initialization stage, a trajectory $\mathcal{T}$ and a zero-padded hidden state $h_\text{init}$ are processed by a bidirectional multi-head self-attention module to obtain context embedding $e^c$ and initial hidden state $h_0$. At each timestep after initialization, the current observation $o_t$ and goal frame $g$ are encoded into $e^s_{t}$ and are taken by a multi-head cross-attention module together with fixed $e^c$ and recurrently updated $h_t$ to produce policy $\pi_\theta$, Q-value $Q_\omega$, and terminal signal $\delta_\upsilon$. The red circle indicates additional aggregation via fusing element-wise product between features
like VLN-Bert~\cite{VLN-Bert}. 
}
\label{fig:structure} 
\end{figure*}

%% file: sec/4_experiment.tex
\section{Experiments}
\label{sec:experiments}
\subsection{Experimental Setups}
\label{sec:Experimental Setups}
\noindent\textbf{Configurations.} 
We apply our algorithm to realistic embodied navigation testbeds RoboTHOR~\cite{Robothor} and Habitat~\cite{Habitat}. For each navigation task in RoboTHOR and Habitat, we initialize a mobile robot with a random position, orientation, and a given goal image extracted from the context video indicating a specific goal object. At each timestep, the agent, mounted with only one RGB camera of $224 \times 224$ resolution, receives an egocentric image and can execute a discrete action from a fixed action space $[{\mathrm{MoveForward, TurnLeft, TurnRight, STOP}}]$. The minimum units of rotation and forward movement are 30\textdegree \ and 0.25 meters respectively. When the goal object is observed and the distance between the goal object and the agent is within 1m, the agent succeeds, otherwise it fails when the maximum step is reached, which is 500 in all experiments. 

\noindent\textbf{Task Division.} 
In RoboTHOR, there are 15 different rooms and 5 possible layouts for each room, in total 75 different scenes. Each room has a different topology from other rooms, and the location of objects in each layout is different from other layouts. In order to test the in-context learning ability at different levels, we divide the training and test scenes as follows. For the training set, we select 12 rooms, and 4 layouts for each room. For the \textbf{unseen layout} testing set, we select the remaining 1 layout of the 12 training set rooms, in total 12 scenes. For the \textbf{unseen room} testing set, we select the remaining 3 rooms and all their layouts, in total 15 scenes.
In Habitat, we divide all 90 building-scale scenes from Matterport3D (MP3D) ~\cite{Matterport3D} into $70$ training scenes and $20$ testing scenes to test the generalization abilities. See Appendix \ref{appendix:dataset} for details.

\noindent\textbf{Dataset Construction.} 
 We use a fixed non-explorative rule-based policy to collect the video dataset. The policy keeps moving forward until colliding with an obstacle and then randomly turns to get rid of it. Simultaneously, the agent keeps recording an egocentric RGB sequence till reaching the maximum 900 steps, \ie a 30-second video clip in each scene. After that, we employ a pretrained object detector Detic~\cite{Detic} to detect all the objects occurred in the video. To get a set for goal frames, we additionally roll out in the simulation to collect goal images of each target object from different views.

\noindent\textbf{Task Evaluation.} 
To evaluate the comprehensive ability of the in-context navigation policy, we design three levels of generalization test:
\begin{itemize}
    \item Evaluate in the \textbf{unseen layout} testing set to test the generalization ability over \textbf{layouts}.
    \item Evaluate in the \textbf{unseen room} testing set to test the generalization ability over \textbf{rooms}. 
    \item Evaluate in scenes from a different simulator to test the generalization ability over \textbf{domains}.
\end{itemize}
For evaluation, the agent is arranged to navigate to each target object detected in a context video. Each navigation task repeats 10 times. 
 
We conduct 3 runs with different random seeds and quantitatively assess the performance using metrics including success rate (SR), success weighted by normalized inverse path length (SPL), trajectory length (TL), and navigation error (NE) following a recent work~\cite{NaVid}.


\noindent\textbf{Baselines.} 
NOLO stands as the first method to incorporate context videos directly into policy for in-context navigation tasks.
To address in-context generalization capabilities, we compare NOLO with a random exploration policy, as well as two large multimodal models (LMMs), GPT-4 and Video-LLaVA which also exhibit remarkable zero-shot multimodal understanding and generation abilities, in a training-free manner. To adapt the LMMs for video-based navigation, we use a queue of size $K$ to store context video frames, current observations, and goal images, which are concatenated as LMM input. A unified prompt guides the LMMs to understand the context and navigate effectively in both RoboTHOR and Habitat, extracting actions from their language responses. See Appendix \ref{appendix:LMM} for details.

In addition, we further compare NOLO with traditional visual navigation methods without context video input, specifically, an online method VGM~\cite{kwon2021visual}, which builds a memory graph based on unsupervised image representations from navigation history to guide actions, and an offline method ZSON~\cite{Zson}, which encodes goal images into a multimodal, semantic embedding space to enable scalable training of SemanticNav agents in unannotated 3D environments. 
Besides, we compare NOLO with a recent work AVDC~\citep{AVDC}, which infers actions from video predictions while incorporating extra depth information. Our experimental results show that AVDC underperforms NOLO even with additional depth images.  
 

\subsection{RoboTHOR}
We average the metrics of video navigation tasks in all testing scenes and showcase the mean results in \cref{tab:performance_comparison}. NOLO demonstrates overwhelming advantages over baselines, including random, LMM baselines, and visual navigation baselines in both testing sets, which proves its strong generalization abilities to new scenes in different generalization levels. The set of unseen rooms is naturally a harder testing set than the unseen layouts in terms of generalization ability. 
Therefore, NOLO displays slightly better navigation performance in the unseen layout testing set (71.92\% SR, 29.26\% SPL) compared to the unseen room testing set (70.48\% SR, 27.74\% SPL). GPT-4o outperforms the random exploration policy across most scenarios, while Video-LLaVA generally performs below the random exploration policy due to its smaller model size relative to GPT-4o. NOLO is also much more parameter-efficient than GPT-4o. However, it achieves better navigation results, underscoring its effectiveness in training in-context navigation policies. Overall, visual navigation baselines tend to surpass both random and LMM baselines because of additional training. However, AVDC exhibits instability in unseen environments due to its reliance on a diffusion model trained on a limited set of egocentric video frames, which leads to inconsistency in video prediction. While VGM’s memory graph and ZSON’s multimodal semantic embeddings can provide some guidance, they still fall short of NOLO in terms of success rate and path efficiency. We analyzed that VGM's implicit graph struggles to effectively represent topological structures with limited data coverage, and for ZSON, there is a non-negligible distribution discrepancy between CLIP~\cite{CLIP} pre-training data and our egocentric videos. See \cref{fig:robothor-radar,fig:robothor-radar-visual} in Appendix \ref{appendix:Results} for detailed results.

\input{tabs/baseline}
\input{figs/cross/cross}
\input{tabs/ablation}
\input{figs/real-maze/real-maze-6tasks1}
\subsection{Habitat}
We evaluate NOLO similarly in Habitat to demonstrate the effectiveness of NOLO in larger-scale scenes. As illustrated in \cref{tab:performance_comparison}
, NOLO still outperforms baselines in terms of SR and achieves slightly better performance in terms of SPL. Compared with the metrics of NOLO in RoboTHOR, the average SPL and SR are lower, and the average NE and TL are higher. This is mostly because scenes in Habitat are significantly larger than those in RoboTHOR, resulting in more challenging navigation tasks across multiple rooms and even floors and requiring longer traversal distances. However, we observe that NOLO is still capable of acquiring many necessary sub-skills like entering or exiting a room, avoiding obstacles, and exploring. See Appendix \ref{appendix:Results} for detailed results.

\subsection{Cross-Domain Evaluation}
To pursue the ultimate goal of video navigation, we evaluate NOLO across RoboTHOR and Habitat, even if the visual inputs may suffer from a severe domain gap. Specifically, NOLO trained from 48 RoboTHOR scenes is evaluated in 20 Habitat scenes \textbf{(R2H)}, and NOLO trained from 70 Habitat scenes is evaluated in 27 RoboTHOR scenes \textbf{(H2R)}. 
For comparison, we also plot the average in-domain results (\textbf{R2R} and \textbf{H2H}). 
We average all the results on testing scenes in \cref{cross}, from which we surprisingly discover that R2H and H2R show competitive performance compared to R2R and H2H. This is because the pretrained image encoder in NOLO provides a general representation of indoor scenes, and the temporal coherence loss in \cref{eq:temporal} also leads to a general representation reflecting the natural temporal ordering, which together lead to the generalization ability of NOLO across domains.

\subsection{Ablation Studies}
\label{sec:ablation}

In this section, we use a set of ablation studies to examine the contribution of several key components in NOLO. More specifically, we aim to answer the following questions.

\noindent\textbf{Does the context video provide informative cues for navigation?} \textbf{Yes.} We conduct an ablation study denoted as \textbf{NOLO-C}, in which we evaluate NOLO without context videos in RoboTHOR and Habitat with other configurations unchanged.
 From \cref{tab:ablation}, a mean performance drop of NOLO-C is perceived both in RoboTHOR and Habitat compared with NOLO due to lack of context information.  

\noindent\textbf{Does temporal coherence loss contribute to better visual representation and performance? } \textbf{Yes.} We examine the effect of the temporal coherence loss in \cref{eq:temporal}. An ablation variant, named \textbf{NOLO-T}, is conducted by removing the temporal coherence loss. The results are shown in \cref{tab:ablation}. The performance drops slightly in RoboTHOR while more obviously in Habitat. To figure out the underlying reasons for their discrepancy in performance, we further compare the learned visual representation of NOLO and NOLO-T. We randomly select a context video in RoboTHOR and utilize t-SNE projection to visualize their embedding sequences encoded by $E_\zeta(f_t)$ in \cref{fig:embedding-sequence} in Appendix \ref{appendix:add-vis}. The embeddings learned by NOLO-T exhibit more irregular stochasticity. In contrast, NOLO learns a better-structured embedding, which demonstrates the effectiveness of the temporal coherence loss.
See Appendix \ref{appendix:Results} for details.

\noindent\textbf{Can NOLO incorporate other off-the-shelf modules to decode actions from videos?} \textbf{Yes.} We conduct an ablation, entitled \textbf{NOLO(M)}, to decode actions from SuperGlue~\cite{Superglue} via point matching instead of GMFlow~\cite{Gmflow}. We observe an accuracy drop in action decoding (from 92.44\% to 80.43\%).
We then follow the same pipeline and report the performance in \cref{tab:ablation}.
NOLO(M) exhibits around 5\% decay in average success rate in RoboTHOR and 7\% in Habitat. Therefore, to achieve more satisfactory in-context generalization, we choose GMFlow as the action decoder for NOLO. However, as a general two-stage framework for in-context learning from videos, NOLO is not limited to any kind of pseudo action labeling methods and is capable of incorporating more advanced modules for better performance.

\noindent\textbf{Does NOLO understand the relationship between goal and context video?} \textbf{Yes}. We use Grad-CAM \cite{CAM} saliency maps of the visual encoder to illustrate NOLO’s ability to connect the goal with the context video. As shown in \cref{fig:goal-context}, when a goal image is provided, NOLO accurately identifies key semantic cues in the goal and focuses on related areas in the context frame, indicating its capacity to link goal and context. Without the goal image, however, NOLO lacks this focus, underscoring its reliance on goal-driven context understanding. This result demonstrates NOLO’s ability to interpret the goal and capture relevant information in the context video.

\subsection{Real-World Experiments}
\label{sec:real}

We constructed a real-world maze environment, as shown in \cref{fig:real-maze}. 
We employed the Unitree Go2 robot, equipped with a RealSense D435i camera to capture RGB observations.
In the maze environment, the robot's actions are discretized into moving forward by 15 cm or turning 30 degrees at each step.
Six target objects--a book, box, cola can, cup, glue, and novel--are placed in the maze. Images of each target are taken from random nearby perspectives.
A fixed, non-exploratory, rule-based policy was employed to collect two 600-step video clips in the maze, one for training and the other as context video during policy deployment.
For evaluation, each episode consists of 100 steps, with a success radius set at 50 cm.
\cref{fig:real-maze-6tasks} illustrates key frames from successful trajectories during in-context policy deployment. The robot demonstrated impressive capabilities in navigating straight passages, executing turns at corners, and ultimately reaching the goals by comprehending the dynamics and topological structure inferred from the unseen context video and its relationship to the goal images.

%% file: tabs/baseline.tex
\begin{table*}[!htbp]\centering 
    \small 
    \caption{Performance comparison of success rates (SR), success path lengths (SPL), trajectory lengths (TL), and navigation errors (NE) across different methods and environments. 
    The blue part shows some training-free methods and the green part shows visual navigation approaches which require training.
    }
    \vspace{-2mm}
    \setlength{\abovecaptionskip}{-10pt} 
    \setlength{\belowcaptionskip}{-10pt} 
    \resizebox{\textwidth}{!}{
    \begin{tabular}{lcccccccccccc}
        \toprule
        \multirow{4}{*}{\textbf{Method}} & \multicolumn{8}{c}{\textbf{Robothor}} & \multicolumn{4}{c}{\textbf{Habitat}} \\
        \cmidrule(lr){2-9} \cmidrule(lr){10-13}
        & \multicolumn{4}{c}{\textbf{Unseen Layout}} & \multicolumn{4}{c}{\textbf{Unseen Room}} & \multicolumn{4}{c}{\textbf{Unseen Room}} \\
        \cmidrule(lr){2-5} \cmidrule(lr){6-9} \cmidrule(lr){10-13}
        & \textbf{SR(\%)$\uparrow$}   & \textbf{SPL(\%)$\uparrow$}   & \textbf{TL$\downarrow$}   & \textbf{NE$\downarrow$} & \textbf{SR(\%)$\uparrow$}   & \textbf{SPL(\%)$\uparrow$}   & \textbf{TL$\downarrow$}   & \textbf{NE$\downarrow$} & \textbf{SR(\%)$\uparrow$}   & \textbf{SPL(\%)$\uparrow$}   & \textbf{TL$\downarrow$}   & \textbf{NE$\downarrow$} \\
        \midrule
        \rowcolor[HTML]{ECF4FF} 
        Random & 26.00 & 13.27 & 394.77 & 2.93 & 23.10 & 12.02 & 405.87 & 2.89 & 24.53 & 10.03 & 402.02 & 4.97 \\
        \rowcolor[HTML]{ECF4FF} 
        GPT-4o & 31.97 & 17.18 & 363.23 & 3.03 & 31.69 & 16.49 & 364.63 & 2.89 & 35.16 & 20.39 & \textbf{345.84} & 4.70 \\
        \rowcolor[HTML]{ECF4FF} 
         Video-LLaVA & 20.07 & 8.31 & 381.92 & 3.08 & 17.51 & 6.58 & 385.72 & 3.32 & 14.55 & 9.02 & 384.53 & 4.71 \\
        \midrule
        \rowcolor[HTML]{E3FAE3}
        AVDC & 32.53 & 9.43 & 430.77 & 2.15 & 31.14 & 11.57 & 424.56 & 2.13 & 27.29 & 12.24 & 418.12 & 4.30 \\
        \rowcolor[HTML]{E3FAE3}
        VGM & 47.04 & 23.90 & 376.92 & 2.44 & 40.51 & 19.99 & 400.90 & 2.38 & 34.34 & 17.30 & 393.92 & 4.19 \\
        \rowcolor[HTML]{E3FAE3}
        ZSON & 37.83 & 21.31 & 349.31 & 2.23 & 35.78 & 18.81 & 372.68 & 2.37 & 32.83 & 19.86 & 395.94 & 4.32 \\
        \midrule
        \textbf{NOLO (ours)} & \textbf{71.92} & \textbf{29.26} & \textbf{238.92} & \textbf{1.79} & \textbf{70.48} & \textbf{27.74} & \textbf{248.44} & \textbf{1.87} & \textbf{43.65} & \textbf{20.77} & 347.57 & \textbf{3.67} \\
        \bottomrule
    \end{tabular}}
    \label{tab:performance_comparison}
\end{table*}

%% file: figs/cross/cross.tex
\begin{figure*}[htbp]
\centering
\setlength{\textfloatsep}{5pt} 
\setlength{\belowcaptionskip}{-4pt}  
\begin{subfigure}{0.235\linewidth}\setlength{\abovecaptionskip}{2pt}
    \centering
    \includegraphics[height=\textwidth]{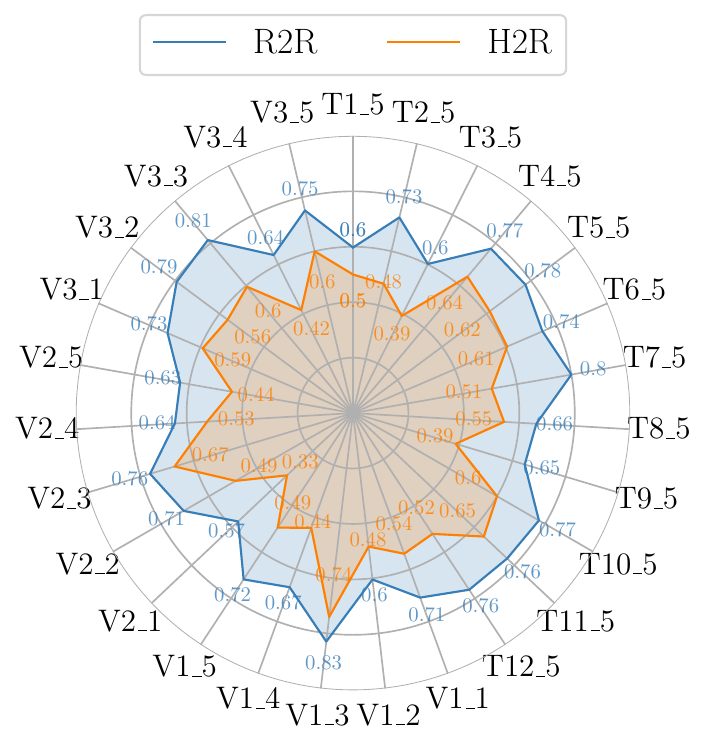}
    \caption{H2R-SR}
    \label{subfig:robothor-sr}
\end{subfigure}
\begin{subfigure}{0.235\linewidth}\setlength{\abovecaptionskip}{2pt}
    \centering
    \includegraphics[height=\textwidth]{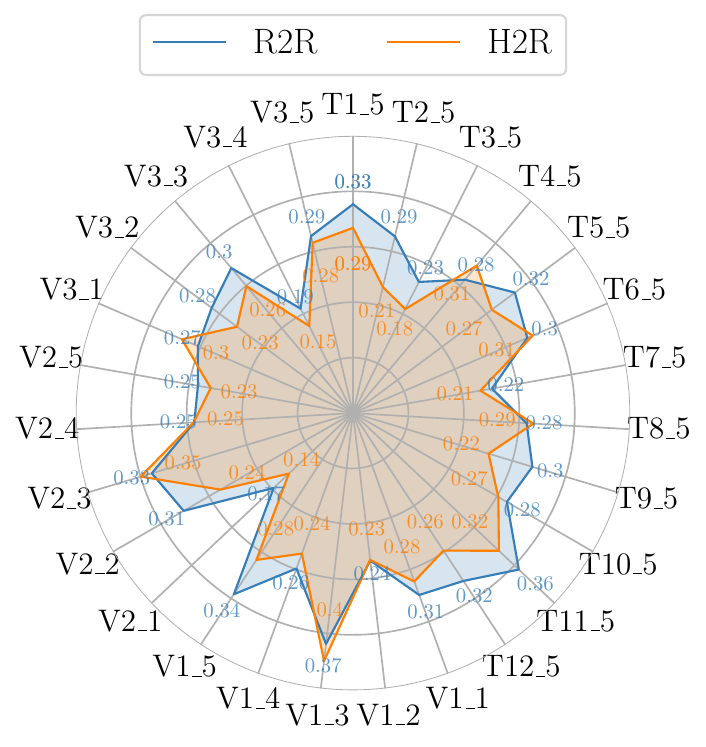}
    \caption{H2R-SPL}
    \label{subfig:robothor-spl}
\end{subfigure}
\begin{subfigure}{0.235\linewidth}\setlength{\abovecaptionskip}{2pt}
    \centering
    \includegraphics[height=\textwidth]{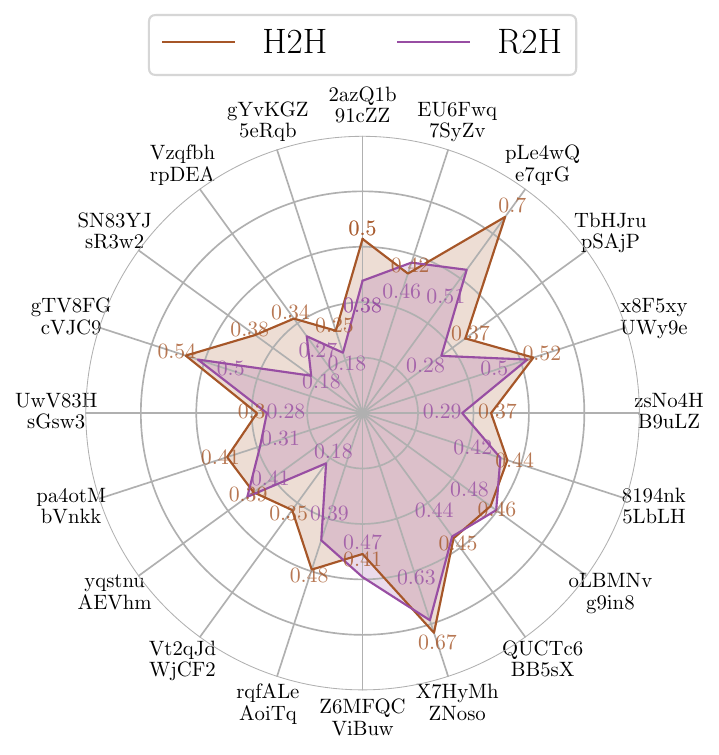}
    \caption{R2H-SR}
    \label{subfig:habitat-sr}
\end{subfigure}
\begin{subfigure}{0.235\linewidth}\setlength{\abovecaptionskip}{2pt}
    \centering
    \includegraphics[height=\textwidth]{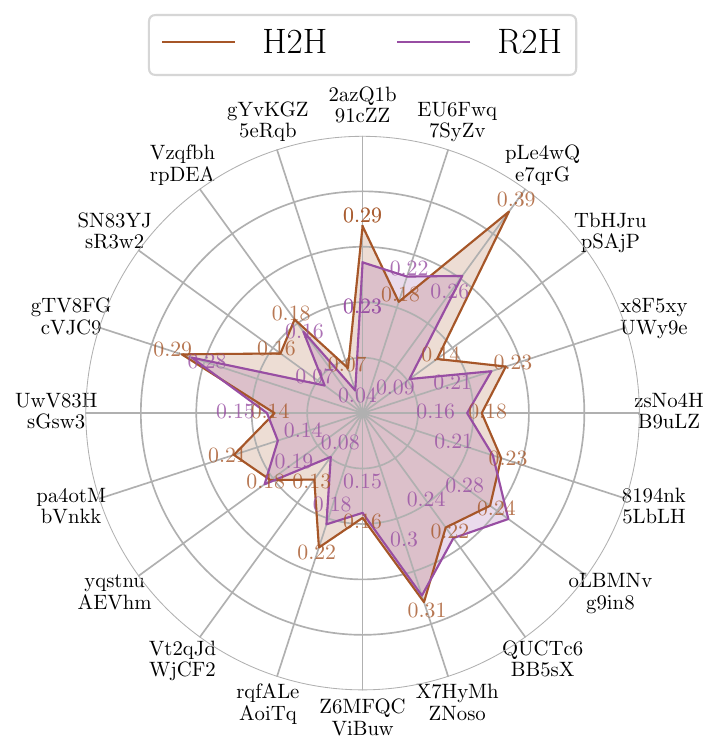}
    \caption{R2H-SPL}
    \label{subfig:habitat-spl}
\end{subfigure}
\caption{Average SR and SPL of cross domain evaluation. H2R means training in Habitat and testing in RoboTHOR, and vice versa.
}
\vspace{-5pt}
\label{cross}
\end{figure*}

%% file: tabs/ablation.tex
\begin{table*}[htbp]\centering 
\caption{Ablation studies of NOLO. For each method, we average the metrics over all testing scenes and average all the mean results over three random seeds to report the mean and standard deviation.}
\vspace{-2mm}
\resizebox{\textwidth}{!}{
\begin{tabular}{@{}lcccccccc@{}}
\toprule
\multirow{2.5}{*}{\textbf{Method}} & \multicolumn{4}{c}{\textbf{Robothor}} & \multicolumn{4}{c}{\textbf{Habitat}} \\ \cmidrule(lr){2-5} \cmidrule(lr){6-9}
                        & \textbf{SR(\%)$\uparrow$}   & \textbf{SPL(\%)$\uparrow$}   & \textbf{TL$\downarrow$}   & \textbf{NE$\downarrow$}  & \textbf{SR(\%)$\uparrow$}   & \textbf{SPL(\%)$\uparrow$}   & \textbf{TL$\downarrow$}  & \textbf{NE$\downarrow$}  \\ \midrule
NOLO & \textbf{71.20} $\pm$ \footnotesize{0.85} & \textbf{28.50} $\pm$ \footnotesize{0.72} & \textbf{243.68} $\pm$ \footnotesize{3.80} & \textbf{1.83} $\pm$ \footnotesize{0.04} & \textbf{43.65} $\pm$ \footnotesize{1.09} & \textbf{20.77} $\pm$ \footnotesize{0.46} & \textbf{347.57} $\pm$ \footnotesize{3.32} & \textbf{3.67} $\pm$ \footnotesize{0.09} \\
NOLO-C & 67.78 $\pm$ \footnotesize{0.60} & 27.70 $\pm$ \footnotesize{1.35} & 259.65 $\pm$ \footnotesize{0.53} & 1.98 $\pm$ \footnotesize{0.02} & 33.58 $\pm$ \footnotesize{0.92} & 17.41 $\pm$ \footnotesize{0.43} & 381.26 $\pm$ \footnotesize{3.23} & 4.58 $\pm$ \footnotesize{0.26} \\
NOLO-T & 69.03 $\pm$ \footnotesize{0.87} & 28.15 $\pm$ \footnotesize{0.09} & 252.84 $\pm$ \footnotesize{6.33} & 1.93 $\pm$ \footnotesize{0.04} & 37.48 $\pm$ \footnotesize{1.01} & 18.51 $\pm$ \footnotesize{0.71} & 369.85 $\pm$ \footnotesize{2.72} & 4.72 $\pm$ \footnotesize{0.07} \\
NOLO(M) & 66.15 $\pm$ \footnotesize{0.37} & 26.66 $\pm$ \footnotesize{1.61} & 267.95 $\pm$ \footnotesize{1.96} & 2.02 $\pm$ \footnotesize{0.00} & 36.73 $\pm$ \footnotesize{1.89} & 18.17 $\pm$ \footnotesize{0.31} & 373.92 $\pm$ \footnotesize{3.46} & 4.69 $\pm$ \footnotesize{0.25} \\ \bottomrule
\end{tabular}
}
\label{tab:ablation}
\end{table*}

%% file: figs/real-maze/real-maze-6tasks1.tex
\begin{figure*}[htbp]
\centering
\includegraphics[width=.76\textwidth]{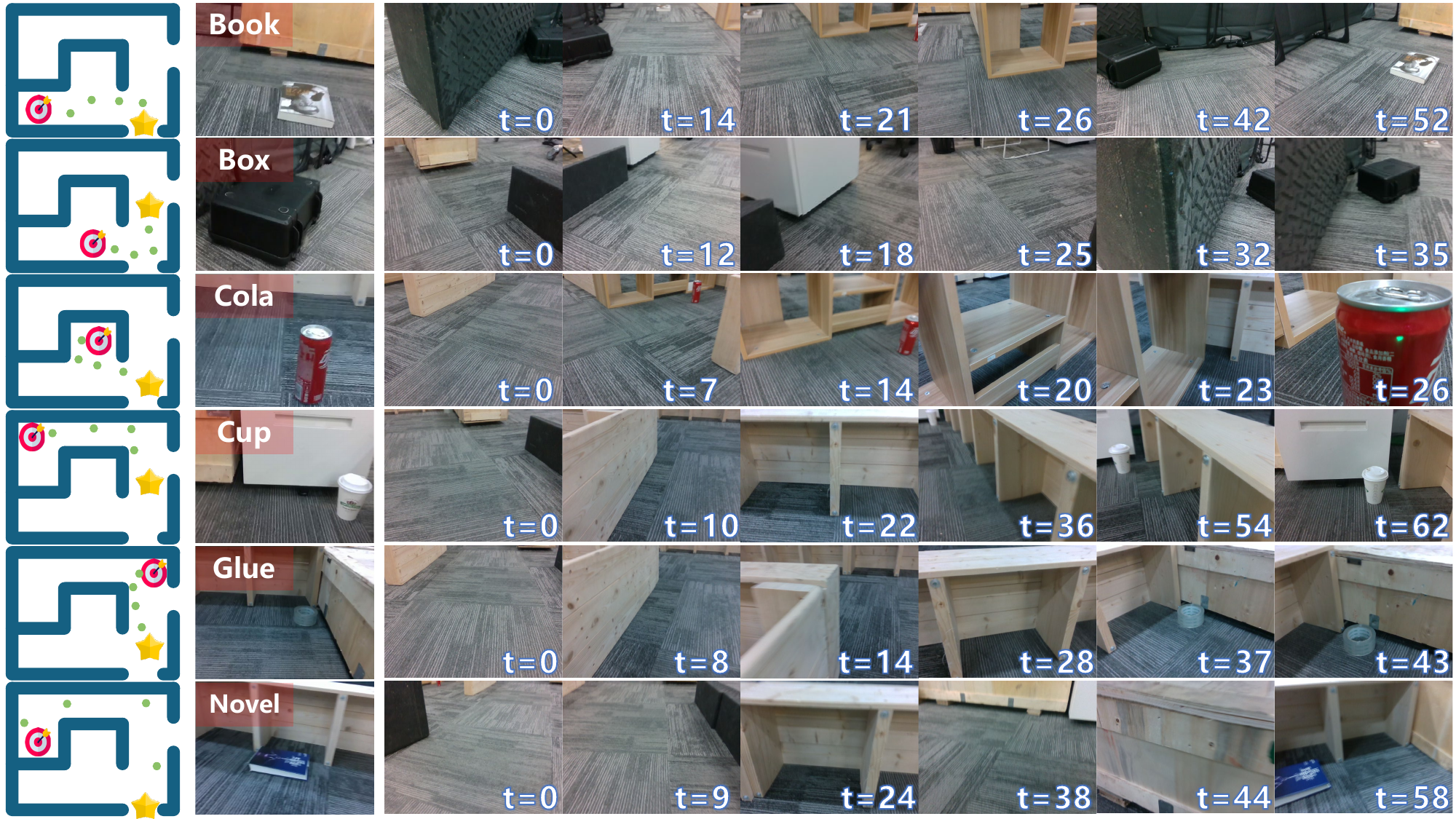}
\setlength{\belowcaptionskip}{-7pt} 
\caption{Visualizations of six video navigation tasks during in-context policy deployment. The leftmost bird-eye-view topological maps depicts the robot positions corresponding to the key observations to the right, and the second column displays the goal images.
}
\label{fig:real-maze-6tasks} 
\end{figure*}

%% file: sec/5_conclusion.tex
\section{Conclusion and Limitaiton}
\label{sec:conclusion}
\input{figs/vis/goal-context1}

\input{figs/real-maze/real-maze-setup1}
In this paper, we propose and investigate the video navigation problem, in which the goal is to offline train a generalizable in-context navigation policy from videos to find objects occurred in a context video about a novel scene. To solve this problem, we propose NOLO, to train a \model \  taking a context video, current observation, and goal frame as input and outputting navigation actions via pseudo action labeling and offline reinforcement learning. Additionally, we introduce a temporal coherence loss to ensure representations align with natural temporal order. Experiments in RoboTHOR and Habitat simulation environments as well as in a real-world maze demonstrate NOLO can generalize to varying novel scenes without finetuning or re-training by understanding the context video. Ablation studies further reveal the impact of NOLO’s key components. 
It is important to emphasize the limitation of NOLO that currently it can only extract movement actions from two adjacent frames. Future work will explore methods for decoding more complex actions from multiple frames. We also aim to pretrain NOLO on larger-scale scene datasets to assess its generalization across more diverse environments.

%% file: figs/vis/goal-context1.tex
\begin{figure}[h]
\centering
\includegraphics[scale=0.37]
{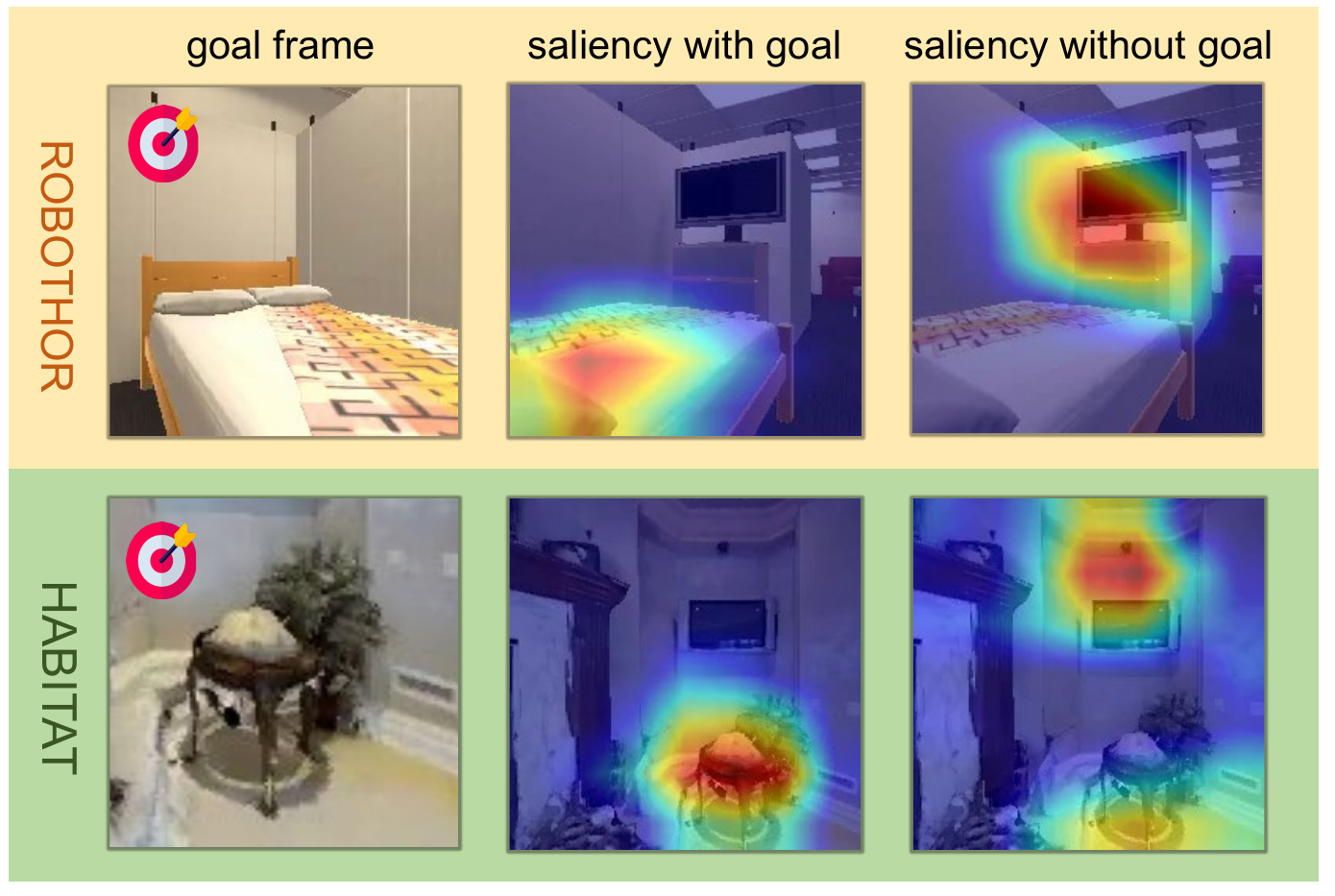}
\caption{Comparision of saliency maps between NOLO and NOLO-C about context frames with respect to goal frames in RoboTHOR and Habitat.  
}
\label{fig:goal-context} 
\end{figure}

%% file: figs/real-maze/real-maze-setup1.tex
\begin{figure}[h]
\centering
\includegraphics[width=.4\textwidth]{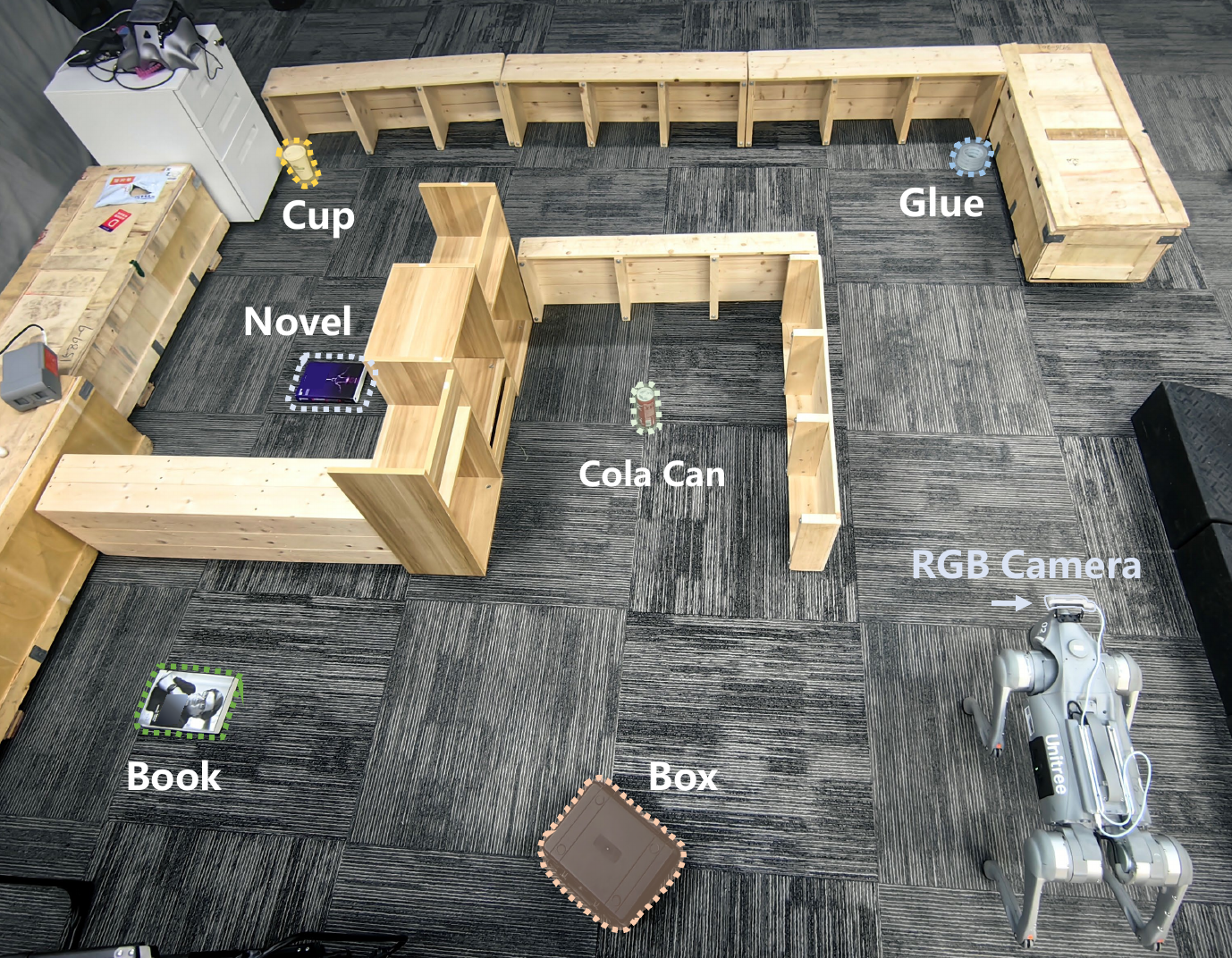}
\caption{Setups of real-world maze. The robotic dog mounted with a head camera acts as the navigation agent and six annotated objects are goals.  
}
\label{fig:real-maze} 
\end{figure}

%% file: sec/X_suppl.tex
\clearpage
\setcounter{page}{1}
\maketitlesupplementary

\section{Methodology Details}
\subsection{Pseudo Action Labeling}
\label{appendix:Pseudo}
We deploy optical flow and filter dominant vectors with the help of the flow map. Theoretically, discretizing filtered vectors into several action categories can be achieved by any unsupervised pattern recognition methods, such as clustering. In practice, we use a simple threshold-based heuristic. We observe that the feature of moving forward is less prominent than turning left or right. Consequently, we focus more on the horizontal displacement of the pixels in the flow map. If the horizontal component of horizontal displacement exceeds a threshold $\tau_x$ and the vertical component stays within a threshold $\tau_y$ then we classify it as a turning action, or it is considered as a forwarding action.   

Occasionally, we notice that the displacement of pixels observed across frames manifests irregular patterns. To mitigate this issue, for each pair of adjacent frames $(f_t,f_{t+1})$, we filter dominant vectors $\nu_t$ that exhibit gradient amplitudes within the upper decile, which are subsequently discretized into actions derived from the action space, making it possible to trace the trajectory of the agent over time. 

\subsection{Policy Inference}
\label{appendix:policy}
We choose the threshold of choosing action $\beta=0.5$ .After the loss of BCQ converges, we make decisions based on the learned policy and Q-values. During inference, an action is chosen based on the Q-values weighted by masked action distribution $\pi _{\theta}\left( a_t|g, o_{t},\mathcal{T}^i \right)$ with probability $\varepsilon=0.999$ and otherwise randomly. Besides, as long as $P_D$ outputs a strong signal for termination, the agent will directly output action $STOP$.

\subsection{Semantic Actions}
To ensure more stable navigation, the {\model} is elaborately designed to output "semantic actions", i.e. actions from the action space along with their duration. For example, $(\mathrm{MoveForward},3)$ means repeatedly moving forward for 3 time steps. $\mathrm{STOP}$ can executed at most once. We find that limiting each action up to \textbf{3} times at most effectively improves the robustness of the learned policy and the final performance. Previous methods~\cite{NRNS,VLV} tend to adopt a more complicated hierarchical framework consisting of a high-level policy to propose sub-goals from a constructed topological map and a low-level policy to decompose several actions in the action space to reach each sub-goal. Instead, we adopt a simpler but more practical mid-level policy to regularize the navigation behavior. From another perspective, we enlarge the original action space to an extended one with 10 new actions. 

\subsection{Model Architecture}
\label{appendix:architecture}
We deploy ResNet18 pretrained from Places368 \cite{Places365} as visual Encoder and frozen ResNet50-based Clip visual encoder as goal image encoder. In {\model}, the self-attention modules consist of 9 Bert \cite{Bert} layers and the cross-attention modules consist of 4 LXMERT \cite{Lxmert} layers. The structure of {\model} is modified from Prevalent
\cite{Prevalent}. The self-attention modules and cross-attention modules are also initialized with the checkpoints of Prevalent. Our codes and documents are included in supplementary materials.

\section{Experiment Details}
\label{appendix:Experiment Details}
\subsection{Dataset Division}
\label{appendix:dataset}
Derived from Ai2THOR
, valid scenes in RoboTHOR
are in the form FloorPlan\_Train\{1:12\}\_\{1:5\} or FloorPlan\_Val\{1:3\}\_\{1:5\}. 
See \cref{tab:scene-split} for training and testing scene splits. We include detailed steps in the README.md contained in the supplementary materials indicating how to run scripts to create video datasets and gather testing points.

\input{tabs/scenelist}

\subsection{LMM Baselines}
\label{appendix:LMM}
For both GPT-4o and Video-LLaVA, we maintain a queue of size $K=16$ to incorporate sampled context frames, current observation, and goal image into a video clip and designed a consistent prompt to facilitate their understanding of the video content and to guide them to provide a reasonable action at each timestep. The prompt is structured as follows:

\begin{quote}
\texttt{These frames pertain to a navigation task. The goal of the video is represented by the last frame, and the second-to-last frame is the current observation. Choose the best action from [move forward, turn left, turn right, stop] that will help achieve the goal depicted in the last frame of the video. Please provide the selected action directly.}
\end{quote}

We then utilize regular expressions to extract actions from the received language responses. In case of invalid output, we attempt up to three retries, and a random action is selected as output if all attempts fail.

\subsection{Hyperparameters}
\label{appendix:hyperparameters}
\cref{tab:hyper} outlines the hyperparameters for {\model} training.

\input{tabs/hyperparameter}

\subsection{Computation Resources}
\label{appendix:resources}
For both RoboTHOR and Habitat, we train NOLO and all NOLO-variants on a single 40G A100 GPU for around 1 day. We evaluate NOLO across all created testing tasks (single thread) on a 24G Nvidia RTX 4090 Ti GPU for about 6 hours in RoboTHOR and 5 hours in Habitat.   

\section{Additional Visualizations}
\label{appendix:add-vis}
\input{figs/vis/emb-seq1}
\cref{appendix:add-vis} compares the t-SNE visual embedding sequence of a context video in RoboTHOR between NOLO and NOLO-T. 

\section{Evaluation Results}
\label{appendix:Results}
Tables below outline the average performance of different methods in RoboTHOR and Habitat as well as cross evaluation results over 3 random seeds.
\subsection{RoboTHOR Evaluation Performance}
We also examined the impact of NOLO compared to LMM baselines and visual navigation baselines on two key metrics for vision-based navigation tasks: Success Rate (SR) and Success weighted by normalized inverse Path Length (SPL), as illustrated in the figures in \cref{fig:robothor-radar} and \cref{fig:robothor-radar-visual}. \cref{subfig:robothor-train-sr,subfig:robothor-train-spl,subfig:robothor-train-sr-visual,subfig:robothor-train-spl-visual} show the performance in the unseen layout testing set. \cref{subfig:robothor-val-sr,subfig:robothor-val-spl,subfig:robothor-val-sr-visual,subfig:robothor-val-spl-visual} show the evaluation results in the unseen room testing set, and The detailed performances of NOLO, random policy, GPT-4o, Video-LLaVA, AVDC, VGM, ZSON, NOLO-C, NOLO-T, NOLO(M) in RoboTHOR are reported in \cref{tab:robothor-nolo-appendix,tab:robothor-random-appendix,tab:robothor-gpt-appendix,tab:robothor-videollava-appendix,tab:robothor-AVDC-appendix,tab:robothor-VGM-appendix,tab:robothor-ZSON-appendix,tab:robothor-nolo-c-appendix,tab:robothor-nolo-t-appendix,tab:robothor-nolom-appendix} respectively. \input{figs/robothor/robothor}
\input{tabs/robothor},

\subsection{Habitat Evaluation Performance}
Also, we examined the impact of NOLO compared to LMM baselines and visual navigation baselines on two key metrics for vision-based navigation tasks: Success Rate (SR) and Success weighted by normalized inverse Path Length (SPL), as illustrated in \cref{fig:habitat-radar-visual}. The performances of NOLO, random policy, GPT-4o, Video-LLaVA, AVDC, VGM, ZSON, NOLO-C, NOLO-T, NOLO(M) in Habitat testing scenes are reported in \cref{tab:habitat-nolo-appendix,tab:habitat-random-appendix,tab:habitat-gpt-appendix,tab:habitat-videollava-appendix,tab:habitat-avdc-appendix,tab:habitat-vgm-appendix,tab:habitat-zson-appendix,tab:habitat-nolo-c-appendix,tab:habitat-nolo-t-appendix,tab:habitat-nolom-appendix} respectively.
\input{figs/habitat/habitat}
\input{tabs/habitat}

\clearpage
\subsection{Cross Evaluation Results}
The performances of H2R and R2H are reported in \cref{tab:rbothor-habitat-appendix,tab:habitat-rbothor-appendix} respectively.
\input{tabs/cross}

\section{Social Impact and Future Work}
\label{appendix: impact}
We see our work as a starting point to propose a general and scalable method to offline learn a navigation policy from limited sources for any mobile agents. In the future, We believe NOLO will benefit a lot from the rapid development in optical flow and computer vision foundation models, and there will be extensive applications of NOLO, such as deployment to sweeping robots or rescue robots. Further improvement can be conducted to examine and enhance the reliability of the decoded actions from optical flow. It is also a great idea to incorporate first-person and third-person videos to showcase more advanced intelligence of spatial understanding and egocentric decision-making in more complex 3D embodied environments. 

%% file: tabs/scenelist.tex
\begin{table} 
\caption{Dataset division of RoboTHOR and Habitat}
\begin{tabular}{lll}
\toprule
\multicolumn{3}{c}{\textbf{Train}} \\ 
\midrule
S9hNv5qa7GM & XcA2TqTSSAj & PX4nDJXEHrG \\  5LpN3gDmAk7 & WYY7iVyf5p8 & 5q7pvUzZiYa \\
VFuaQ6m2Qom & EDJbREhghzL & rPc6DW4iMge \\
2n8kARJN3HM & HxpKQynjfin & PuKPg4mmafe \\
759xd9YjKW5 & aayBHfsNo7d & jtcxE69GiFV \\
B6ByNegPMKs & p5wJjkQkbXX & D7N2EKCX4Sj \\
1LXtFkjw3qL & YVUC4YcDtcY & YmJkqBEsHnH \\
7y3sRwLe3Va & r47D5H71a5s & vyrNrziPKCB \\
gxdoqLR6rwA & 1pXnuDYAj8r & VLzqgDo317F \\
kEZ7cmS4wCh & sT4fr6TAbpF & ARNzJeq3xxb \\
Vvot9Ly1tCj & cV4RVeZvu5T & VVfe2KiqLaN \\
jh4fc5c5qoQ & 2t7WUuJeko7 & D7G3Y4RVNrH \\
uNb9QFRL6hY & V2XKFyX4ASd & e9zR4mvMWw7 \\
RPmz2sHmrrY & E9uDoFAP3SH & 17DRP5sb8fy \\
JeFG25nYj2p & mJXqzFtmKg4 & YFuZgdQ5vWj \\
ac26ZMwG7aT & JF19kD82Mey & Pm6F8kyY3z2 \\
sKLMLpTHeUy & wc2JMjhGNzB & dhjEzFoUFzH \\
82sE5b5pLXE & gZ6f7yhEvPG & 8WUmhLawc2A \\
5ZKStnWn8Zo & pRbA3pwrgk9 & 29hnd4uzFmX \\
GdvgFV5R1Z5 & ULsKaCPVFJR & q9vSo1VnCiC \\
Uxmj2M2itWa & qoiz87JEwZ2 & s8pcmisQ38h \\
ur6pFq6Qu1A & fzynW3qQPVF & ZMojNkEp431 \\
r1Q1Z4BcV1o & b8cTxDM8gDG & i5noydFURQK \\
JmbYfDe2QKZ &  &  \\ 
\midrule
\multicolumn{3}{c}{\textbf{Eval}} \\ \cmidrule[0.75pt]{1-3}
2azQ1b91cZZ & x8F5xyUWy9e & QUCTc6BB5sX \\
Vt2qJdWjCF2 & gTV8FGcVJC9 & EU6Fwq7SyZv \\
zsNo4HB9uLZ & X7HyMhZNoso & yqstnuAEVhm \\
SN83YJsR3w2 & pLe4wQe7qrG & 8194nk5LbLH \\
Z6MFQCViBuw & pa4otMbVnkk & VzqfbhrpDEA \\
TbHJrupSAjP & oLBMNvg9in8 & rqfALeAoiTq \\
UwV83HsGsw3 & gYvKGZ5eRqb & \\ 
\bottomrule
\end{tabular}
\label{tab:scene-split}
\end{table}

%% file: tabs/hyperparameter.tex
\begin{table}[htbp]\centering 
\caption{Hyperparameters for {\model} Training}
~\\
\begin{tabular}{@{}ll@{}} 
\toprule
\textbf{Hyperparameter} & \textbf{Value}\\ \midrule
Optimizer & AdamW \\
Learning Rate & 1e-4 \\
Betas & (0.9,0.95)\\
Weight Decay & 0.1\\
Batch size & 26\\
SA Layers & 9 \\
CA Layers & 4 \\
Action Embedding Dimension & 256\\
Visual Embedding Dimension & 512\\
Hidden State Dimension & 768\\
BCQ Clip Threshold & 0.5\\
Type of GPUs & A100 \& Nvidia RTX 4090 Ti \\
\bottomrule
\end{tabular}
\label{tab:hyper}
\end{table}

%% file: figs/vis/emb-seq1.tex

\begin{figure}[htbp]
\centering
\includegraphics[width=.48\textwidth]{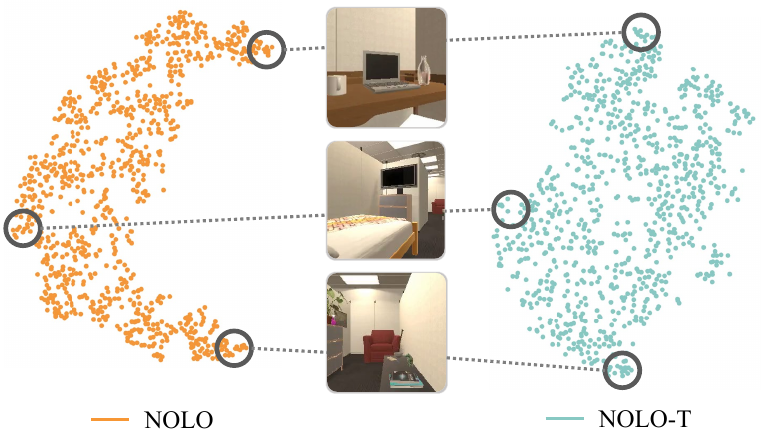}
\caption{Comparision of context embedding sequences of a sampled video from RoboTHOR between NOLO and NOLO-T. 
}
\label{fig:embedding-sequence} 
\end{figure}

%% file: figs/robothor/robothor.tex
\begin{figure*}[htbp]
\centering

\begin{subfigure}[t]{0.99\textwidth}\setlength{\abovecaptionskip}{0.1pt}
    \centering
    \includegraphics[scale=0.35]{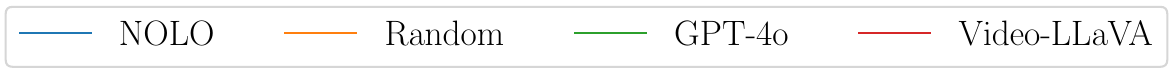}
\end{subfigure}

\begin{subfigure}{0.21\linewidth}\setlength{\abovecaptionskip}{2pt}
    \centering
    \includegraphics[height=\textwidth]{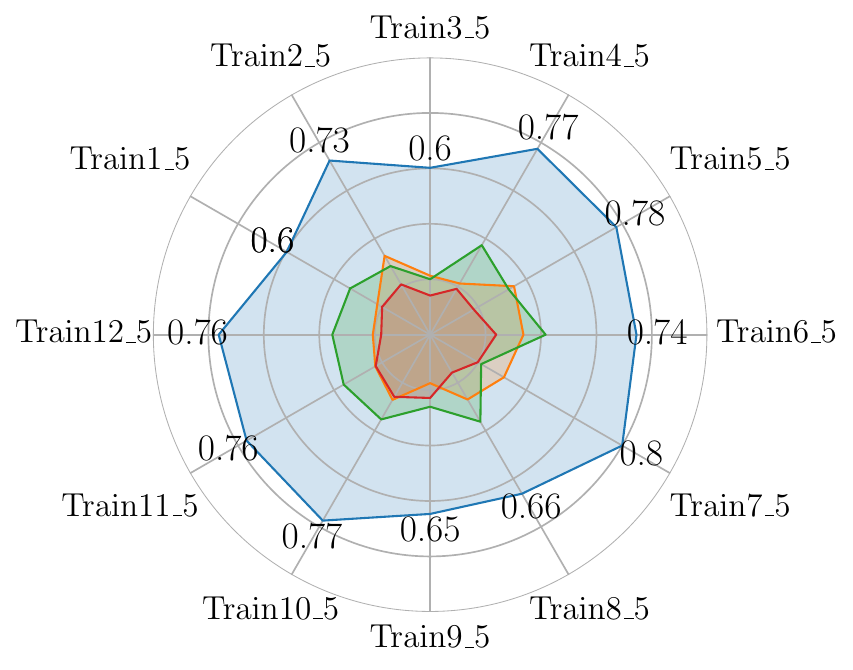}
    \caption{Unseen Layout-SR}
    \label{subfig:robothor-train-sr}
\end{subfigure}
\hspace{7mm}
\begin{subfigure}{0.21\linewidth}\setlength{\abovecaptionskip}{2pt}
    \centering
    \includegraphics[height=\textwidth]{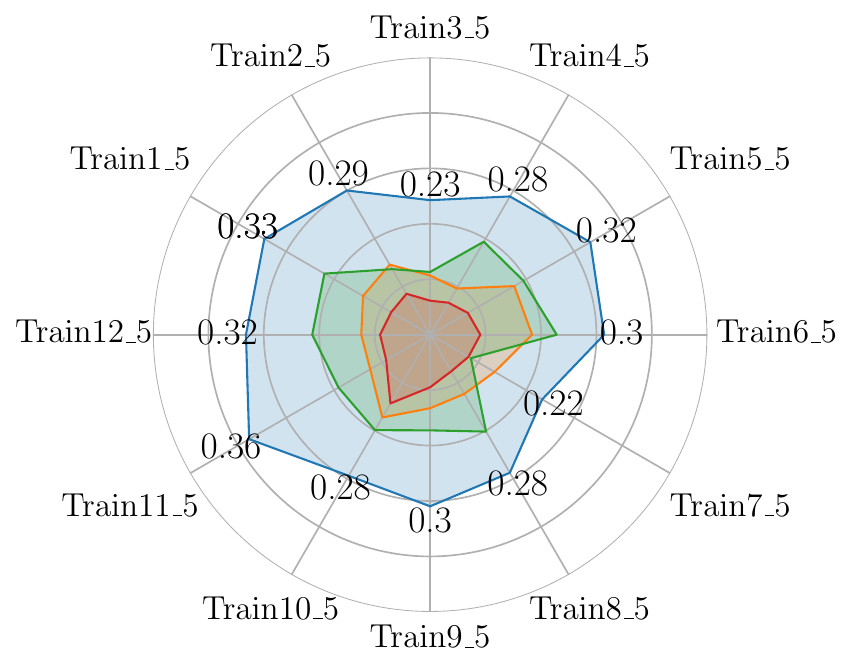}
    \caption{Unseen Layout-SPL}
    \label{subfig:robothor-train-spl}
\end{subfigure}
\hspace{7mm}
\begin{subfigure}{0.21\linewidth}\setlength{\abovecaptionskip}{2pt}
    \centering
    \includegraphics[height=\textwidth]{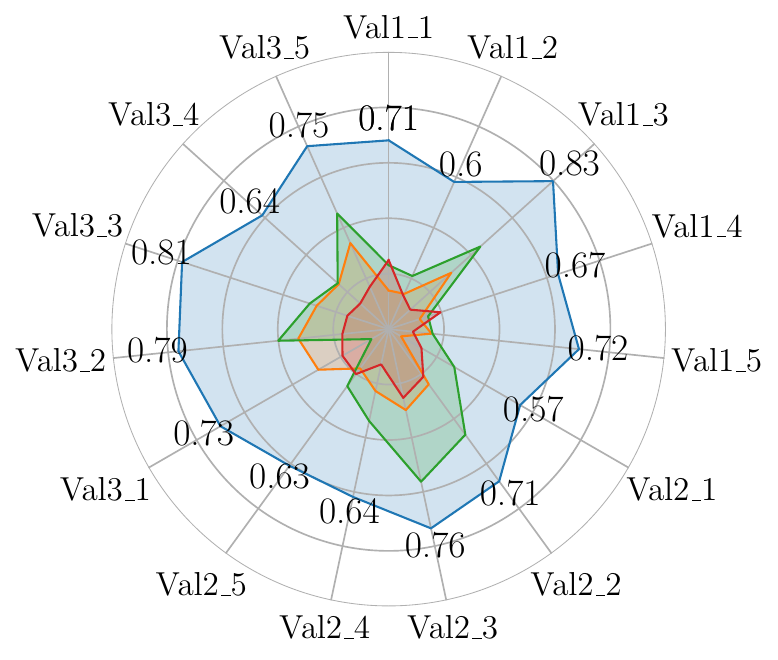}
    \caption{Unseen Room-SR}
    \label{subfig:robothor-val-sr}
\end{subfigure}
\hspace{3mm}
\begin{subfigure}{0.21\linewidth}\setlength{\abovecaptionskip}{2pt}
    \centering
    \includegraphics[height=\textwidth]{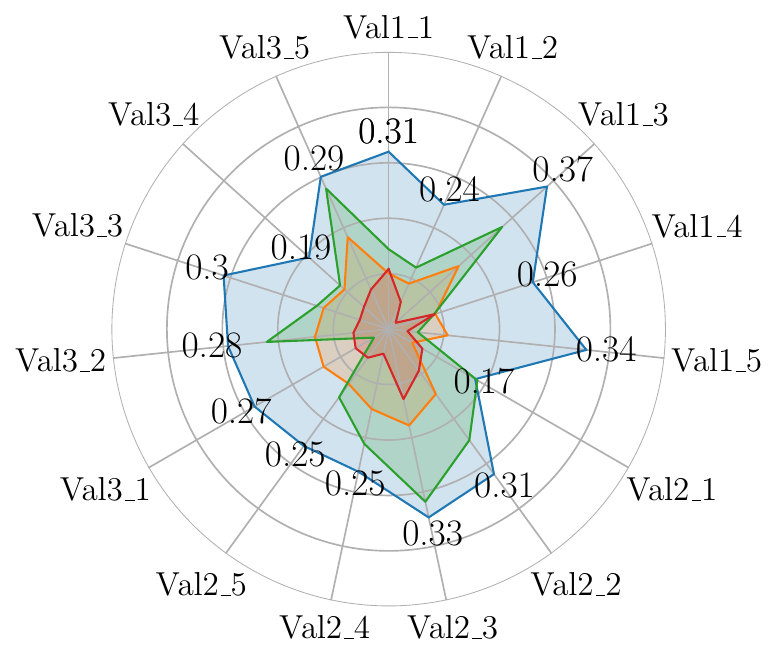}
    \caption{Unseen Room-SPL}
    \label{subfig:robothor-val-spl}
\end{subfigure}
\caption{Compare NOLO with random policy and LMM baselines in RoboTHOR seen and unseen rooms.}
\label{fig:robothor-radar}
\end{figure*}

\begin{figure*}[htbp]
\centering

\begin{subfigure}[t]{0.99\textwidth}\setlength{\abovecaptionskip}{0.1pt}
    \centering
    \includegraphics[scale=0.35]{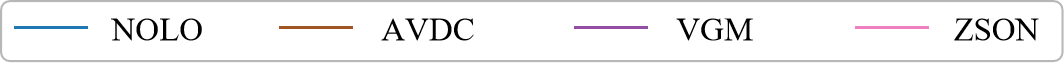}
\end{subfigure}

\begin{subfigure}{0.21\linewidth}\setlength{\abovecaptionskip}{2pt}
    \centering
    \includegraphics[height=\textwidth]{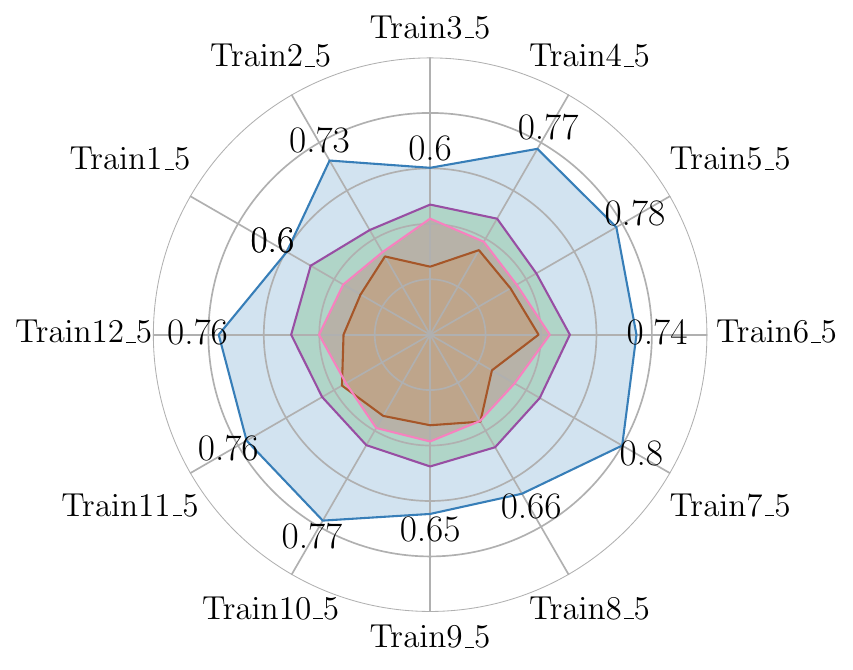}
    \caption{Unseen Layout-SR}
    \label{subfig:robothor-train-sr-visual}
\end{subfigure}
\hspace{7mm}
\begin{subfigure}{0.21\linewidth}\setlength{\abovecaptionskip}{2pt}
    \centering
    \includegraphics[height=\textwidth]{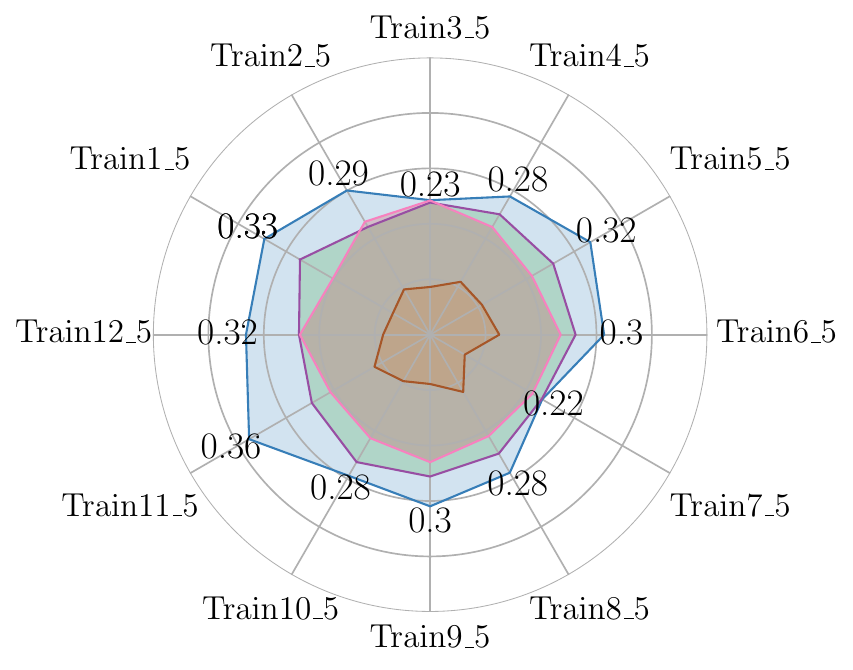}
    \caption{Unseen Layout-SPL}
    \label{subfig:robothor-train-spl-visual}
\end{subfigure}
\hspace{7mm}
\begin{subfigure}{0.21\linewidth}\setlength{\abovecaptionskip}{2pt}
    \centering
    \includegraphics[height=\textwidth]{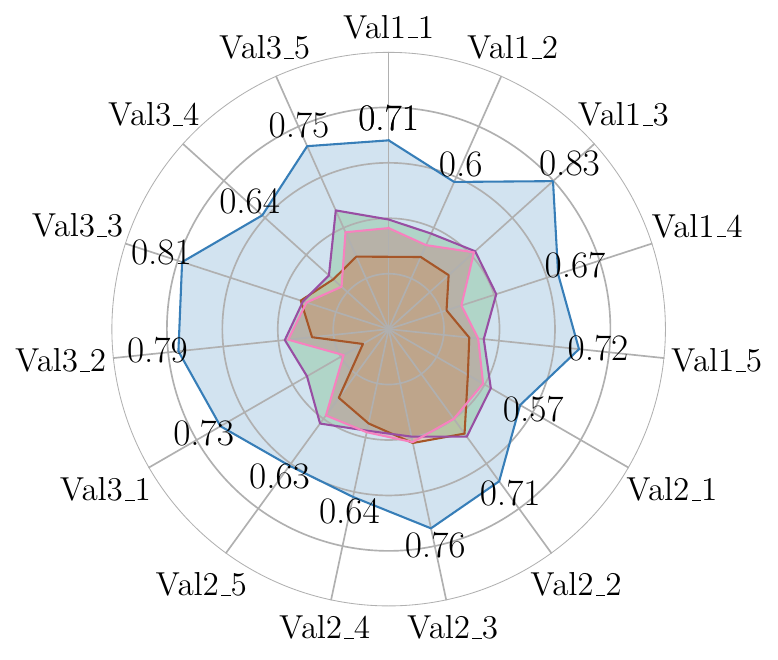}
    \caption{Unseen Room-SR}
    \label{subfig:robothor-val-sr-visual}
\end{subfigure}
\hspace{3mm}
\begin{subfigure}{0.21\linewidth}\setlength{\abovecaptionskip}{2pt}
    \centering
    \includegraphics[height=\textwidth]{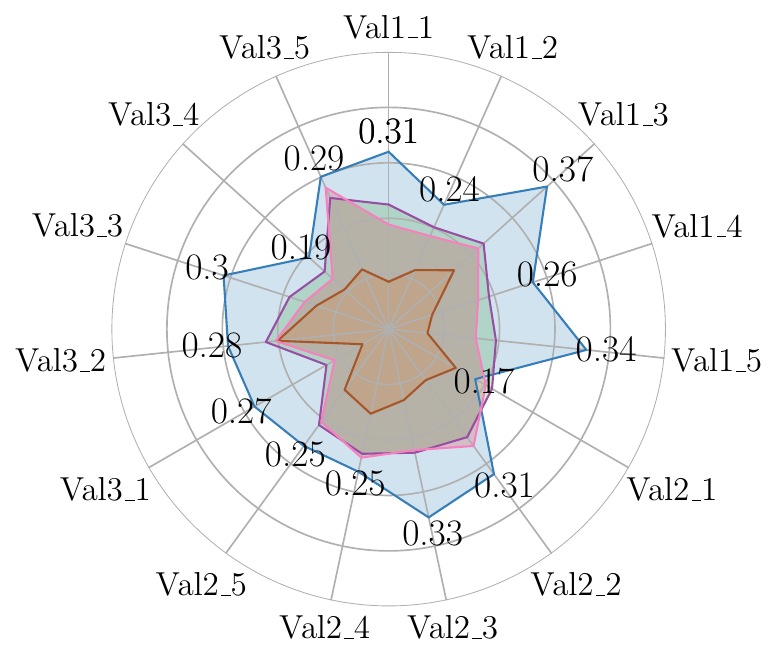}
    \caption{Unseen Room-SPL}
    \label{subfig:robothor-val-spl-visual}
\end{subfigure}
\caption{Compare NOLO with visual navigation baselines in RoboTHOR seen and unseen rooms.}
\label{fig:robothor-radar-visual}
\end{figure*}

%% file: tabs/robothor.tex
\begin{table}[htbp]\centering 
\caption{NOLO performance on RoboTHOR testing set. ``T'' denotes FloorPlan\_Train and ``V'' denotes FloorPlan\_Val.}
\label{tab:robothor-nolo-appendix}
\resizebox{0.5\textwidth}{!}{
\begin{tabular}{@{}lcccc@{}} 
\toprule
\textbf{Scene} & \textbf{SR(\%)$\uparrow$} & \textbf{SPL(\%)$\uparrow$} & \textbf{TL$\downarrow$} & \textbf{NE$\downarrow$}\\ \midrule

T10\_5 & 77.45$\pm$ \footnotesize{4.36} & 28.21$\pm$ \footnotesize{1.58} & 216.35$\pm$ \footnotesize{14.21} & 1.68$\pm$ \footnotesize{0.13} \\
T11\_5 & 76.44$\pm$ \footnotesize{3.82} & 36.19$\pm$ \footnotesize{2.32} & 218.66$\pm$ \footnotesize{13.06} & 1.62$\pm$ \footnotesize{0.16} \\
T12\_5 & 76.27$\pm$ \footnotesize{3.54} & 31.89$\pm$ \footnotesize{0.94} & 228.36$\pm$ \footnotesize{4.96} & 1.50$\pm$ \footnotesize{0.09} \\
T1\_5 & 59.72$\pm$ \footnotesize{2.39} & 33.15$\pm$ \footnotesize{1.72} & 260.69$\pm$ \footnotesize{9.97} & 2.06$\pm$ \footnotesize{0.05} \\
T2\_5 & 72.54$\pm$ \footnotesize{0.81} & 28.83$\pm$ \footnotesize{1.45} & 231.64$\pm$ \footnotesize{4.56} & 1.76$\pm$ \footnotesize{0.11} \\
T3\_5 & 60.20$\pm$ \footnotesize{1.21} & 23.29$\pm$ \footnotesize{0.21} & 289.70$\pm$ \footnotesize{6.56} & 2.34$\pm$ \footnotesize{0.04} \\
T4\_5 & 77.42$\pm$ \footnotesize{2.74} & 27.64$\pm$ \footnotesize{1.66} & 216.64$\pm$ \footnotesize{6.83} & 1.71$\pm$ \footnotesize{0.13} \\
\bottomrule
\end{tabular}
}
\end{table}


\begin{table}[htbp]\centering 
\caption{Random performance on RoboTHOR testing set. ``T'' denotes FloorPlan\_Train and ``V'' denotes FloorPlan\_Val.}
\label{tab:robothor-random-appendix}
\resizebox{0.5\textwidth}{!}{
\begin{tabular}{@{}lcccc@{}} 
\toprule
\textbf{Scene} & \textbf{SR(\%)$\uparrow$} & \textbf{SPL(\%)$\uparrow$} & \textbf{TL$\downarrow$} & \textbf{NE$\downarrow$}\\ \midrule

T10\_5 & 27.15$\pm$ \footnotesize{0.50} & 16.55$\pm$ \footnotesize{0.85} & 371.96$\pm$ \footnotesize{3.07} & 2.84$\pm$ \footnotesize{0.00} \\
T11\_5 & 22.83$\pm$ \footnotesize{4.50} & 11.98$\pm$ \footnotesize{2.67} & 406.66$\pm$ \footnotesize{15.35} & 3.24$\pm$ \footnotesize{0.22} \\
T12\_5 & 20.68$\pm$ \footnotesize{1.26} & 11.93$\pm$ \footnotesize{0.51} & 413.58$\pm$ \footnotesize{4.25} & 3.02$\pm$ \footnotesize{0.04} \\
T1\_5 & 22.00$\pm$ \footnotesize{2.00} & 13.41$\pm$ \footnotesize{1.47} & 396.63$\pm$ \footnotesize{0.20} & 3.08$\pm$ \footnotesize{0.09} \\
T2\_5 & 32.83$\pm$ \footnotesize{0.93} & 14.00$\pm$ \footnotesize{0.57} & 369.18$\pm$ \footnotesize{2.40} & 2.54$\pm$ \footnotesize{0.06} \\
T3\_5 & 21.26$\pm$ \footnotesize{6.38} & 10.25$\pm$ \footnotesize{0.81} & 421.76$\pm$ \footnotesize{1.69} & 3.36$\pm$ \footnotesize{0.09} \\
T4\_5 & 21.32$\pm$ \footnotesize{2.23} & 9.23$\pm$ \footnotesize{1.62} & 415.40$\pm$ \footnotesize{9.71} & 3.08$\pm$ \footnotesize{0.06} \\
\bottomrule
\end{tabular}
}
\end{table}


\begin{table}[htbp]\centering 
\caption{GPT-4o performance on RoboTHOR testing set. ``T'' denotes FloorPlan\_Train and ``V'' denotes FloorPlan\_Val.}
\label{tab:robothor-gpt-appendix}
\resizebox{0.4\textwidth}{!}{
\begin{tabular}{@{}lcccc@{}} 
\toprule
\textbf{Scene} & \textbf{SR(\%)$\uparrow$} & \textbf{SPL(\%)$\uparrow$} & \textbf{TL$\downarrow$} & \textbf{NE$\downarrow$}\\ \midrule

T10\_5 & 35.29 & 19.09 & 358.81 & 3.00 \\
T11\_5 & 36.00 & 18.32 & 363.55 & 3.15 \\
T12\_5 & 35.29 & 20.42 & 359.38 & 3.17 \\
T1\_5 & 33.33 & 21.14 & 362.02 & 2.78 \\
T2\_5 & 28.57 & 13.10 & 367.66 & 2.76 \\
T3\_5 & 20.00 & 10.86 & 377.68 & 3.30 \\
T4\_5 & 37.27 & 18.58 & 357.89 & 3.12 \\
T5\_5 & 32.73 & 18.66 & 363.37 & 2.95 \\
T6\_5 & 41.67 & 21.92 & 351.87 & 3.24 \\
T7\_5 & 21.25 & 8.14 & 374.45 & 3.10 \\
T8\_5 & 36.25 & 19.38 & 356.12 & 3.07 \\
T9\_5 & 26.00 & 16.57 & 366.02 & 2.72 \\
\midrule
Average & 31.97 & 17.18 & 363.23 & 3.03 \\
\midrule
V1\_1 & 24.00 & 13.84 & 374.19 & 3.45 \\
V1\_2 & 21.82 & 11.66 & 373.75 & 3.22 \\
V1\_3 & 46.32 & 26.44 & 342.01 & 2.53 \\
V1\_4 & 15.56 & 8.23 & 385.98 & 3.48 \\
V1\_5 & 16.59 & 5.07 & 390.28 & 3.10 \\
V2\_1 & 28.43 & 17.75 & 365.67 & 3.31 \\
V2\_2 & 48.95 & 23.77 & 346.67 & 2.09 \\
V2\_3 & 58.52 & 30.56 & 329.43 & 1.82 \\
V2\_4 & 35.23 & 20.31 & 360.92 & 2.09 \\
V2\_5 & 26.41 & 14.60 & 370.46 & 3.04 \\
V3\_1 & 7.48 & 2.94 & 391.37 & 4.18 \\
V3\_2 & 41.67 & 21.22 & 349.08 & 2.99 \\
V3\_3 & 31.11 & 13.08 & 369.33 & 2.73 \\
V3\_4 & 25.71 & 11.31 & 373.28 & 3.22 \\
V3\_5 & 47.50 & 26.64 & 346.99 & 2.09 \\
\midrule
Average & 31.69 & 16.49 & 364.63 & 2.89 \\

\bottomrule
\end{tabular}
}
\end{table}


\begin{table}[htbp]\centering 
\caption{Video-LLaVA performance on RoboTHOR testing set. ``T'' denotes FloorPlan\_Train and ``V'' denotes FloorPlan\_Val.}
\label{tab:robothor-videollava-appendix}
\resizebox{0.4\textwidth}{!}{
\begin{tabular}{@{}lcccc@{}} 
\toprule
\textbf{Scene} & \textbf{SR(\%)$\uparrow$} & \textbf{SPL(\%)$\uparrow$} & \textbf{TL$\downarrow$} & \textbf{NE$\downarrow$}\\ \midrule

T10\_5 & 25.88 & 13.74 & 370.67 & 2.86 \\
T11\_5 & 22.67 & 8.79 & 380.99 & 3.58 \\
T12\_5 & 17.65 & 8.67 & 379.87 & 3.30 \\
T1\_5 & 20.00 & 7.78 & 383.98 & 2.77 \\
T2\_5 & 20.95 & 8.18 & 382.31 & 2.67 \\
T3\_5 & 14.12 & 5.89 & 388.76 & 3.84 \\
T4\_5 & 19.09 & 6.39 & 385.62 & 3.31 \\
T5\_5 & 18.18 & 7.55 & 385.68 & 3.04 \\
T6\_5 & 23.84 & 8.69 & 378.77 & 2.78 \\
T7\_5 & 19.86 & 7.67 & 382.04 & 3.01 \\
T8\_5 & 15.79 & 7.32 & 384.94 & 2.84 \\
T9\_5 & 22.83 & 9.11 & 379.34 & 2.98 \\
\midrule
Average & 20.07 & 8.31 & 381.92 & 3.08 \\
\midrule
V1\_1 & 25.92 & 10.42 & 374.65 & 3.01 \\
V1\_2 & 13.64 & 5.17 & 389.82 & 3.44 \\
V1\_3 & 10.92 & 1.66 & 396.36 & 3.65 \\
V1\_4 & 20.47 & 8.40 & 382.13 & 3.13 \\
V1\_5 & 9.17 & 3.28 & 395.26 & 3.63 \\
V2\_1 & 14.21 & 6.74 & 388.84 & 3.61 \\
V2\_2 & 22.12 & 8.90 & 377.86 & 3.30 \\
V2\_3 & 26.38 & 12.39 & 372.44 & 3.13 \\
V2\_4 & 13.53 & 4.34 & 391.22 & 4.01 \\
V2\_5 & 20.85 & 6.12 & 387.35 & 3.21 \\
V3\_1 & 20.00 & 6.59 & 381.67 & 2.93 \\
V3\_2 & 17.46 & 6.13 & 387.79 & 2.94 \\
V3\_3 & 16.29 & 5.24 & 388.70 & 2.91 \\
V3\_4 & 14.36 & 5.76 & 388.94 & 3.55 \\
V3\_5 & 17.38 & 7.51 & 382.77 & 3.34 \\
\midrule
Average & 17.51 & 6.58 & 385.72 & 3.32 \\

\bottomrule
\end{tabular}
}
\end{table}

\begin{table}[htbp]\centering 
\caption{AVDC performance on RoboTHOR testing set. T'' denotes FloorPlan\_Train and V'' denotes FloorPlan\_Val.}
\label{tab:robothor-AVDC-appendix}
\resizebox{0.4\textwidth}{!}{
\begin{tabular}{@{}lcccc@{}} 
\toprule
\textbf{Scene} & \textbf{SR(\%)$\uparrow$} & \textbf{SPL(\%)$\uparrow$} & \textbf{TL$\downarrow$} & \textbf{NE$\downarrow$}\\ \midrule

T10\_5 & 33.84 & 9.29 & 422.72 & 1.92 \\
T11\_5 & 36.72 & 11.13 & 434.58 & 2.22 \\
T12\_5 & 31.24 & 8.14 & 437.61 & 2.12 \\
T1\_5 & 28.98 & 7.44 & 428.73 & 2.00 \\
T2\_5 & 32.59 & 9.05 & 440.12 & 2.14 \\
T3\_5 & 24.53 & 8.26 & 444.36 & 2.45 \\
T4\_5 & 35.21 & 10.57 & 423.94 & 1.99 \\
T5\_5 & 33.47 & 10.36 & 433.51 & 2.09 \\
T6\_5 & 39.12 & 11.98 & 415.89 & 2.32 \\
T7\_5 & 25.74 & 6.92 & 439.82 & 2.18 \\
T8\_5 & 36.23 & 11.45 & 421.16 & 2.27 \\
T9\_5 & 32.67 & 8.54 & 426.75 & 2.15 \\
\midrule
Average & 32.53 & 9.43 & 430.77 & 2.15 \\
\midrule
V1\_1 & 27.14 & 8.21 & 462.48 & 2.65 \\
V1\_2 & 29.62 & 11.18 & 440.87 & 2.53 \\
V1\_3 & 30.28 & 15.24 & 405.27 & 1.84 \\
V1\_4 & 22.84 & 7.91 & 454.28 & 2.59 \\
V1\_5 & 30.41 & 6.76 & 447.91 & 2.21 \\
V2\_1 & 34.32 & 13.43 & 425.68 & 2.05 \\
V2\_2 & 48.45 & 10.92 & 398.67 & 1.25 \\
V2\_3 & 43.67 & 12.57 & 387.45 & 1.73 \\
V2\_4 & 36.18 & 14.97 & 404.22 & 1.49 \\
V2\_5 & 31.84 & 12.98 & 412.88 & 2.29 \\
V3\_1 & 11.14 & 5.23 & 450.75 & 3.07 \\
V3\_2 & 28.89 & 19.34 & 409.38 & 1.97 \\
V3\_3 & 34.67 & 13.15 & 429.52 & 2.21 \\
V3\_4 & 27.93 & 10.27 & 439.74 & 2.55 \\
V3\_5 & 29.78 & 11.33 & 399.28 & 1.53 \\
\midrule
Average & 31.14 & 11.57 & 424.56 & 2.13 \\

\bottomrule
\end{tabular}
}
\end{table}

\begin{table}[htbp]\centering 
\caption{VGM performance on RoboTHOR testing set. T'' denotes FloorPlan\_Train and V'' denotes FloorPlan\_Val.}
\label{tab:robothor-VGM-appendix}
\resizebox{0.4\textwidth}{!}{
\begin{tabular}{@{}lcccc@{}} 
\toprule
\textbf{Scene} & \textbf{SR(\%)$\uparrow$} & \textbf{SPL(\%)$\uparrow$} & \textbf{TL$\downarrow$} & \textbf{NE$\downarrow$} \\ \midrule

T10\_5 & 46.02 & 25.47 & 375.88 & 2.49 \\
T11\_5 & 44.98 & 23.65 & 392.45 & 2.71 \\
T12\_5 & 50.13 & 22.71 & 378.29 & 2.39 \\
T1\_5 & 49.75 & 26.02 & 371.15 & 2.55 \\
T2\_5 & 43.59 & 21.54 & 388.67 & 2.23 \\
T3\_5 & 46.93 & 22.88 & 367.36 & 2.67 \\
T4\_5 & 48.32 & 24.07 & 381.74 & 2.41 \\
T5\_5 & 44.25 & 24.61 & 373.28 & 2.44 \\
T6\_5 & 50.41 & 25.19 & 369.24 & 2.33 \\
T7\_5 & 45.67 & 22.32 & 365.72 & 2.62 \\
T8\_5 & 46.89 & 23.78 & 380.91 & 2.35 \\
T9\_5 & 47.54 & 24.56 & 378.32 & 2.13 \\
\midrule
Average & 47.04 & 23.90 & 376.92 & 2.44 \\
\midrule
V1\_1 & 41.13 & 21.57 & 405.87 & 2.64 \\
V1\_2 & 39.23 & 19.31 & 412.88 & 2.87 \\
V1\_3 & 43.67 & 22.13 & 386.24 & 1.77 \\
V1\_4 & 42.45 & 18.27 & 416.54 & 2.41 \\
V1\_5 & 35.88 & 18.68 & 401.96 & 2.54 \\
V2\_1 & 44.26 & 20.67 & 382.33 & 2.38 \\
V2\_2 & 49.87 & 23.12 & 394.56 & 1.85 \\
V2\_3 & 41.22 & 21.87 & 385.19 & 1.91 \\
V2\_4 & 38.96 & 22.07 & 396.72 & 2.25 \\
V2\_5 & 43.78 & 20.52 & 405.33 & 2.64 \\
V3\_1 & 35.37 & 12.43 & 420.64 & 3.22 \\
V3\_2 & 39.16 & 21.38 & 389.33 & 2.42 \\
V3\_3 & 33.78 & 18.04 & 406.57 & 2.11 \\
V3\_4 & 30.15 & 14.92 & 416.38 & 2.82 \\
V3\_5 & 48.76 & 24.87 & 392.94 & 1.87 \\
\midrule
Average & 40.51 & 19.99 & 400.90 & 2.38 \\

\bottomrule
\end{tabular}
}
\end{table}

\begin{table}[htbp]\centering 
\caption{ZSON performance on RoboTHOR testing set. T'' denotes FloorPlan\_Train and V'' denotes FloorPlan\_Val.}
\label{tab:robothor-ZSON-appendix}
\resizebox{0.4\textwidth}{!}{
\begin{tabular}{@{}lcccc@{}} 
\toprule
\textbf{Scene} & \textbf{SR(\%)$\uparrow$} & \textbf{SPL(\%)$\uparrow$} & \textbf{TL$\downarrow$} & \textbf{NE$\downarrow$} \\ \midrule

T10\_5 & 38.89 & 20.69 & 350.67 & 2.23 \\
T11\_5 & 35.06 & 19.92 & 362.44 & 2.41 \\
T12\_5 & 40.06 & 22.54 & 345.23 & 2.16 \\
T1\_5 & 36.17 & 19.45 & 358.91 & 2.09 \\
T2\_5 & 34.37 & 22.58 & 351.12 & 2.02 \\
T3\_5 & 41.86 & 23.24 & 340.11 & 2.34 \\
T4\_5 & 38.72 & 21.52 & 346.72 & 2.29 \\
T5\_5 & 36.06 & 20.35 & 349.33 & 2.25 \\
T6\_5 & 43.17 & 22.61 & 335.84 & 2.38 \\
T7\_5 & 35.17 & 20.45 & 360.17 & 2.15 \\
T8\_5 & 35.96 & 20.27 & 348.52 & 2.31 \\
T9\_5 & 38.52 & 22.08 & 342.66 & 2.18 \\
\midrule
Average & 37.83 & 21.31 & 349.31 & 2.23 \\
\midrule
V1\_1 & 37.91 & 18.12 & 378.87 & 2.72 \\
V1\_2 & 34.50 & 17.64 & 386.19 & 2.91 \\
V1\_3 & 43.06 & 20.88 & 359.72 & 1.77 \\
V1\_4 & 28.71 & 16.29 & 395.18 & 2.56 \\
V1\_5 & 33.77 & 15.17 & 380.14 & 2.64 \\
V2\_1 & 40.93 & 19.42 & 361.48 & 2.12 \\
V2\_2 & 41.60 & 25.01 & 353.26 & 1.43 \\
V2\_3 & 43.34 & 21.41 & 368.11 & 1.85 \\
V2\_4 & 39.55 & 22.75 & 354.24 & 2.25 \\
V2\_5 & 40.02 & 19.83 & 360.73 & 2.72 \\
V3\_1 & 19.52 & 10.81 & 392.54 & 3.22 \\
V3\_2 & 37.89 & 19.64 & 368.18 & 2.56 \\
V3\_3 & 32.41 & 15.28 & 382.45 & 2.21 \\
V3\_4 & 23.77 & 13.04 & 390.76 & 2.89 \\
V3\_5 & 39.79 & 26.81 & 358.32 & 1.73 \\
\midrule
Average & 35.78 & 18.81 & 372.68 & 2.37 \\

\bottomrule
\end{tabular}
}
\end{table}

\begin{table}[htbp]\centering 
\caption{NOLO-C performance on RoboTHOR testing set. ``T'' denotes FloorPlan\_Train and ``V'' denotes FloorPlan\_Val.}
\label{tab:robothor-nolo-c-appendix}
\resizebox{0.5\textwidth}{!}{
\begin{tabular}{@{}lcccc@{}} 
\toprule
\textbf{Scene} & \textbf{SR(\%)$\uparrow$} & \textbf{SPL(\%)$\uparrow$} & \textbf{TL$\downarrow$} & \textbf{NE$\downarrow$}\\ \midrule

T10\_5 & 72.94$\pm$ \footnotesize{5.08} & 27.83$\pm$ \footnotesize{0.15} & 252.83$\pm$ \footnotesize{8.62} & 1.86$\pm$ \footnotesize{0.29} \\
T11\_5 & 82.00$\pm$ \footnotesize{1.44} & 36.28$\pm$ \footnotesize{2.90} & 195.77$\pm$ \footnotesize{8.38} & 1.50$\pm$ \footnotesize{0.05} \\
T12\_5 & 73.53$\pm$ \footnotesize{0.83} & 32.13$\pm$ \footnotesize{1.26} & 230.53$\pm$ \footnotesize{8.63} & 1.59$\pm$ \footnotesize{0.05} \\
T1\_5 & 64.72$\pm$ \footnotesize{1.04} & 32.41$\pm$ \footnotesize{1.39} & 258.61$\pm$ \footnotesize{2.70} & 1.94$\pm$ \footnotesize{0.06} \\
T2\_5 & 70.63$\pm$ \footnotesize{3.40} & 29.21$\pm$ \footnotesize{3.18} & 243.38$\pm$ \footnotesize{16.56} & 1.78$\pm$ \footnotesize{0.11} \\
T3\_5 & 62.16$\pm$ \footnotesize{1.94} & 23.55$\pm$ \footnotesize{0.62} & 288.69$\pm$ \footnotesize{8.45} & 2.27$\pm$ \footnotesize{0.09} \\
T4\_5 & 72.12$\pm$ \footnotesize{5.77} & 27.65$\pm$ \footnotesize{0.21} & 244.76$\pm$ \footnotesize{16.37} & 1.89$\pm$ \footnotesize{0.20} \\
\bottomrule
\end{tabular}
}
\end{table}


\begin{table}[htbp]\centering 
\caption{NOLO-T performance on RoboTHOR testing set. ``T'' denotes FloorPlan\_Train and ``V'' denotes FloorPlan\_Val.}
\label{tab:robothor-nolo-t-appendix}
\resizebox{0.5\textwidth}{!}{
\begin{tabular}{@{}lcccc@{}} 
\toprule
\textbf{Scene} & \textbf{SR(\%)$\uparrow$} & \textbf{SPL(\%)$\uparrow$} & \textbf{TL$\downarrow$} & \textbf{NE$\downarrow$}\\ \midrule

T10\_5 & 76.47$\pm$ \footnotesize{3.53} & 29.97$\pm$ \footnotesize{2.31} & 225.76$\pm$ \footnotesize{18.05} & 1.71$\pm$ \footnotesize{0.16} \\
T11\_5 & 78.00$\pm$ \footnotesize{0.67} & 36.00$\pm$ \footnotesize{0.43} & 210.91$\pm$ \footnotesize{7.94} & 1.56$\pm$ \footnotesize{0.10} \\
T12\_5 & 75.88$\pm$ \footnotesize{0.59} & 30.77$\pm$ \footnotesize{1.29} & 233.25$\pm$ \footnotesize{4.82} & 1.62$\pm$ \footnotesize{0.02} \\
T1\_5 & 64.17$\pm$ \footnotesize{2.50} & 32.47$\pm$ \footnotesize{4.30} & 254.97$\pm$ \footnotesize{6.37} & 2.07$\pm$ \footnotesize{0.01} \\
T2\_5 & 65.71$\pm$ \footnotesize{6.67} & 26.12$\pm$ \footnotesize{4.84} & 259.38$\pm$ \footnotesize{29.69} & 1.95$\pm$ \footnotesize{0.25} \\
T3\_5 & 59.41$\pm$ \footnotesize{4.12} & 22.67$\pm$ \footnotesize{0.98} & 303.37$\pm$ \footnotesize{11.79} & 2.39$\pm$ \footnotesize{0.08} \\
T4\_5 & 72.73$\pm$ \footnotesize{0.91} & 27.02$\pm$ \footnotesize{2.10} & 238.05$\pm$ \footnotesize{10.76} & 1.89$\pm$ \footnotesize{0.09} \\
\bottomrule
\end{tabular}
}
\end{table}


\begin{table}[htbp]\centering 
\caption{NOLO(M) performance on RoboTHOR testing set. ``T'' denotes FloorPlan\_Train and ``V'' denotes FloorPlan\_Val.}
\label{tab:robothor-nolom-appendix}
\resizebox{0.5\textwidth}{!}{
\begin{tabular}{@{}lcccc@{}} 
\toprule
\textbf{Scene} & \textbf{SR(\%)$\uparrow$} & \textbf{SPL(\%)$\uparrow$} & \textbf{TL$\downarrow$} & \textbf{NE$\downarrow$}\\ \midrule

T10\_5 & 80.59$\pm$ \footnotesize{4.12} & 32.92$\pm$ \footnotesize{1.93} & 202.68$\pm$ \footnotesize{22.39} & 1.47$\pm$ \footnotesize{0.12} \\
T11\_5 & 77.33$\pm$ \footnotesize{2.67} & 36.00$\pm$ \footnotesize{3.27} & 218.04$\pm$ \footnotesize{1.64} & 1.64$\pm$ \footnotesize{0.01} \\
T12\_5 & 78.82 & 36.62$\pm$ \footnotesize{0.82} & 211.44$\pm$ \footnotesize{9.89} & 1.55$\pm$ \footnotesize{0.01} \\
T1\_5 & 50.83$\pm$ \footnotesize{5.83} & 27.99$\pm$ \footnotesize{2.85} & 302.14$\pm$ \footnotesize{14.69} & 2.35$\pm$ \footnotesize{0.10} \\
T2\_5 & 69.05$\pm$ \footnotesize{1.43} & 25.09$\pm$ \footnotesize{2.83} & 256.49$\pm$ \footnotesize{1.55} & 1.85$\pm$ \footnotesize{0.13} \\
T3\_5 & 54.71$\pm$ \footnotesize{0.59} & 21.49$\pm$ \footnotesize{0.94} & 312.69$\pm$ \footnotesize{3.18} & 2.61$\pm$ \footnotesize{0.08} \\
T4\_5 & 73.64$\pm$ \footnotesize{2.73} & 25.06$\pm$ \footnotesize{1.03} & 245.16$\pm$ \footnotesize{17.39} & 1.78$\pm$ \footnotesize{0.08} \\
\bottomrule
\end{tabular}
}
\end{table}

%% file: figs/habitat/habitat.tex

\begin{figure}[htbp]
\centering

\begin{subfigure}[t]{0.5\textwidth}
    \centering
    \includegraphics[scale=0.35]{./figs/legend.pdf}
\end{subfigure}

\begin{subfigure}[t]{0.22\textwidth}\setlength{\abovecaptionskip}{0.1pt}
    \centering
    \includegraphics[width=\textwidth]{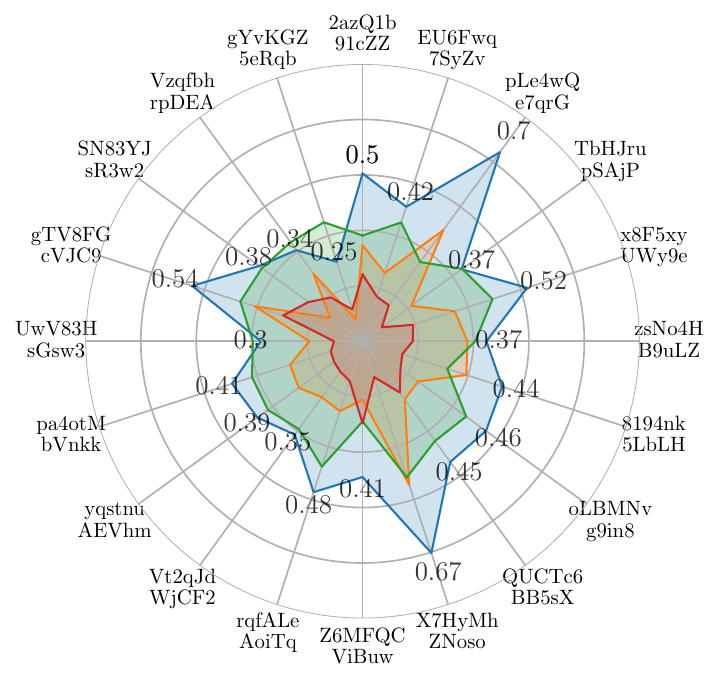}
    \caption{Average SR }
    \label{fig:habitat-sr}
\end{subfigure}
\hspace{1mm}
\begin{subfigure}[t]{0.22\textwidth}\setlength{\abovecaptionskip}{0.1pt}
    \centering
    \includegraphics[width=\textwidth]{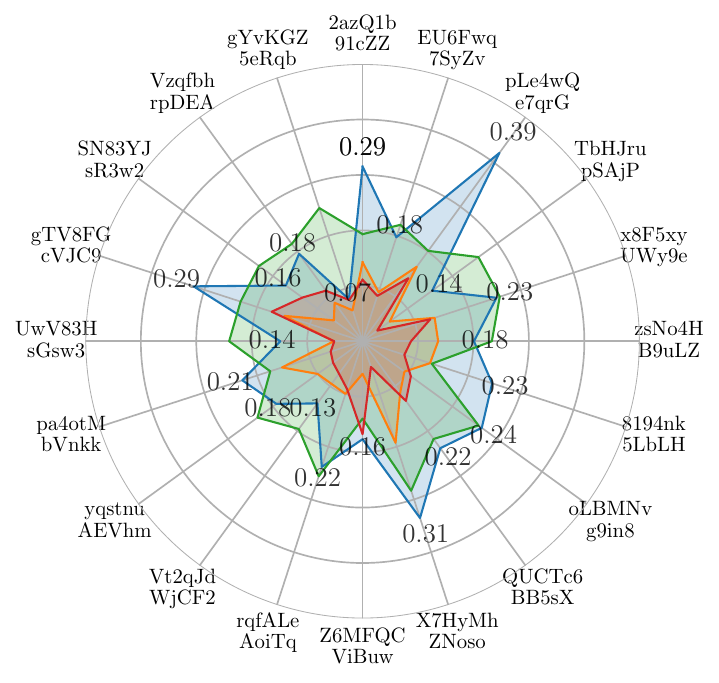}
    \caption{Average SPL }
    \label{fig:habitat-spl}
\end{subfigure}


\begin{subfigure}[t]{0.5\textwidth}
    \centering
    \includegraphics[scale=0.35]{./figs/legend_visual.png}
\end{subfigure}

\begin{subfigure}[t]{0.22\textwidth}\setlength{\abovecaptionskip}{0.1pt}
    \centering
    \includegraphics[width=\textwidth]{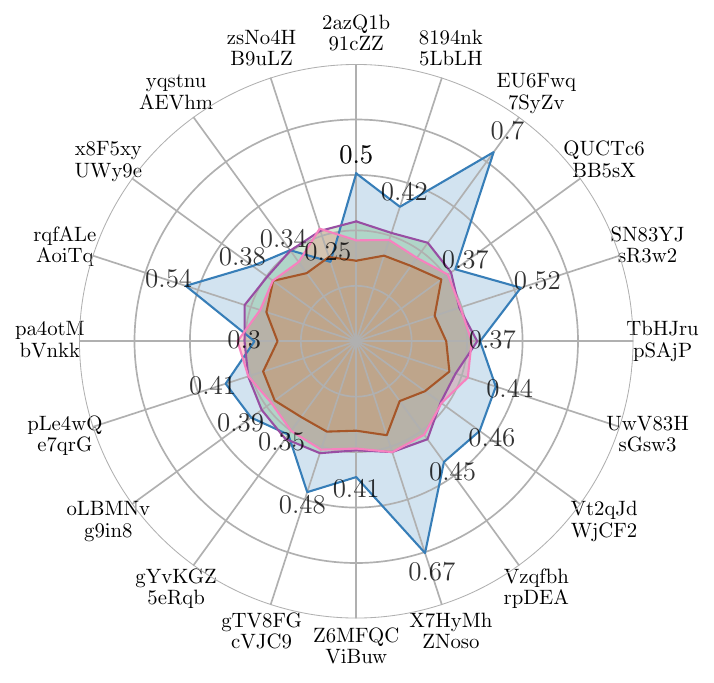}
    \caption{Average SR }
    \label{fig:habitat-sr-visual}
\end{subfigure}
\hspace{1mm}
\begin{subfigure}[t]{0.22\textwidth}\setlength{\abovecaptionskip}{0.1pt}
    \centering
    \includegraphics[width=\textwidth]{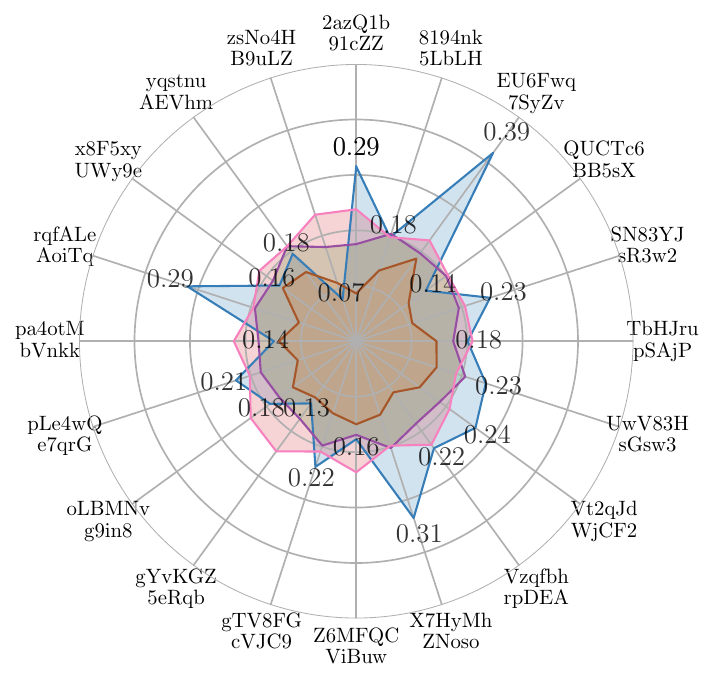}
    \caption{Average SPL }
    \label{fig:habitat-spl-visual}
\end{subfigure}
\caption{
Compare NOLO with random policy, LMM baselines and visual navigation baselines in 20 Habitat testing scenes.
}
\label{fig:habitat-radar-visual}
\end{figure}

%% file: tabs/habitat.tex
\clearpage
\begin{table}[htbp]
\caption{NOLO performance in Habitat testing scenes.}
\label{tab:habitat-nolo-appendix}
\resizebox{0.5\textwidth}{!}{
\begin{tabular}{@{}lcccc@{}} 
\toprule
\textbf{Scene} & \textbf{SR(\%)$\uparrow$} & \textbf{SPL(\%)$\uparrow$} & \textbf{TL$\downarrow$} & \textbf{NE$\downarrow$}\\ \midrule
2azQ1b91cZZ & 50.26$\pm$ \footnotesize{3.10} & 28.99$\pm$ \footnotesize{1.40} & 311.44$\pm$ \footnotesize{11.24} & 2.80$\pm$ \footnotesize{0.12} \\
8194nk5LbLH & 43.94$\pm$ \footnotesize{1.55} & 22.60$\pm$ \footnotesize{0.51} & 341.28$\pm$ \footnotesize{1.28} & 3.10$\pm$ \footnotesize{0.08} \\
EU6Fwq7SyZv & 42.33$\pm$ \footnotesize{1.70} & 18.15$\pm$ \footnotesize{1.13} & 362.99$\pm$ \footnotesize{5.59} & 2.90$\pm$ \footnotesize{0.06} \\
QUCTc6BB5sX & 44.67$\pm$ \footnotesize{6.24} & 21.94$\pm$ \footnotesize{1.36} & 344.75$\pm$ \footnotesize{21.03} & 4.25$\pm$ \footnotesize{0.84} \\
SN83YJsR3w2 & 38.33$\pm$ \footnotesize{6.24} & 15.74$\pm$ \footnotesize{2.48} & 379.77$\pm$ \footnotesize{17.81} & 2.97$\pm$ \footnotesize{0.97} \\
TbHJrupSAjP & 36.67$\pm$ \footnotesize{10.19} & 14.29$\pm$ \footnotesize{3.23} & 376.57$\pm$ \footnotesize{28.90} & 3.85$\pm$ \footnotesize{0.79} \\
UwV83HsGsw3 & 30.48$\pm$ \footnotesize{3.75} & 13.66$\pm$ \footnotesize{0.67} & 402.48$\pm$ \footnotesize{14.42} & 4.80$\pm$ \footnotesize{0.58} \\
\bottomrule
\end{tabular}
}
\end{table}

\begin{table}[htbp]\centering 
\caption{Random performance in Habitat testing scenes.}
\label{tab:habitat-random-appendix}
\resizebox{0.5\textwidth}{!}{
\begin{tabular}{@{}lcccc@{}} 
\toprule
\textbf{Scene} & \textbf{SR(\%)$\uparrow$} & \textbf{SPL(\%)$\uparrow$} & \textbf{TL$\downarrow$} & \textbf{NE$\downarrow$}\\ \midrule

2azQ1b91cZZ & 28.73$\pm$ \footnotesize{6.04} & 13.11$\pm$ \footnotesize{3.30} & 386.34$\pm$ \footnotesize{24.39} & 3.91$\pm$ \footnotesize{0.26} \\
8194nk5LbLH & 32.68$\pm$ \footnotesize{4.95} & 11.77$\pm$ \footnotesize{0.59} & 380.37$\pm$ \footnotesize{10.02} & 3.78$\pm$ \footnotesize{0.13} \\
EU6Fwq7SyZv & 21.50$\pm$ \footnotesize{2.50} & 8.57$\pm$ \footnotesize{0.97} & 409.05$\pm$ \footnotesize{12.59} & 5.07$\pm$ \footnotesize{0.18} \\
QUCTc6BB5sX & 21.50$\pm$ \footnotesize{0.50} & 10.60$\pm$ \footnotesize{1.00} & 399.77$\pm$ \footnotesize{3.10} & 5.67$\pm$ \footnotesize{0.00} \\
SN83YJsR3w2 & 12.00$\pm$ \footnotesize{3.00} & 5.89$\pm$ \footnotesize{1.05} & 428.02$\pm$ \footnotesize{12.97} & 4.97$\pm$ \footnotesize{0.14} \\
TbHJrupSAjP & 18.07$\pm$ \footnotesize{0.21} & 5.57$\pm$ \footnotesize{1.55} & 427.09$\pm$ \footnotesize{10.29} & 5.89$\pm$ \footnotesize{0.37} \\
UwV83HsGsw3 & 15.93$\pm$ \footnotesize{1.93} & 4.78$\pm$ \footnotesize{1.42} & 435.46$\pm$ \footnotesize{1.59} & 5.57$\pm$ \footnotesize{0.16} \\
\bottomrule
\end{tabular}
}
\end{table}

\begin{table}[htbp]\centering 
\caption{GPT-4o performance in Habitat testing scenes.}
\label{tab:habitat-gpt-appendix}
\resizebox{0.5\textwidth}{!}{
\begin{tabular}{@{}lcccc@{}} 
\toprule
\textbf{Scene} & \textbf{SR(\%)$\uparrow$} & \textbf{SPL(\%)$\uparrow$} & \textbf{TL$\downarrow$} & \textbf{NE$\downarrow$}\\ \midrule

2azQ1b91cZZ & 31.67 & 17.77 & 348.79 & 4.19 \\
8194nk5LbLH & 26.67 & 12.07 & 360.59 & 4.21 \\
EU6Fwq7SyZv & 37.41 & 20.31 & 347.59 & 5.10 \\
QUCTc6BB5sX & 36.90 & 20.04 & 347.43 & 6.38 \\
SN83YJsR3w2 & 37.24 & 21.26 & 344.79 & 4.69 \\
TbHJrupSAjP & 36.89 & 23.76 & 339.05 & 2.99 \\
UwV83HsGsw3 & 32.86 & 22.14 & 344.40 & 4.97 \\
Vt2qJdWjCF2 & 32.50 & 18.00 & 350.69 & 7.19 \\
VzqfbhrpDEA & 36.67 & 19.97 & 346.93 & 5.26 \\
X7HyMhZNoso & 42.86 & 26.08 & 335.63 & 2.66 \\
Z6MFQCViBuw & 24.00 & 12.83 & 362.88 & 7.99 \\
gTV8FGcVJC9 & 38.49 & 21.29 & 344.03 & 4.25 \\
gYvKGZ5eRqb & 37.46 & 23.24 & 340.84 & 4.58 \\
oLBMNvg9in8 & 38.34 & 23.79 & 335.41 & 2.06 \\
pLe4wQe7qrG & 29.32 & 18.54 & 350.50 & 5.77 \\
pa4otMbVnkk & 34.82 & 16.11 & 351.36 & 5.98 \\
rqfALeAoiTq & 39.49 & 23.56 & 339.48 & 3.32 \\
x8F5xyUWy9e & 40.91 & 23.94 & 338.39 & 4.23 \\
yqstnuAEVhm & 35.00 & 21.56 & 345.06 & 3.41 \\
zsNo4HB9uLZ & 33.75 & 21.45 & 342.88 & 4.72 \\
\midrule
Average & 35.16 & 20.39 & 345.84 & 4.70 \\
\bottomrule
\end{tabular}
}
\end{table}

\begin{table}[h]
\caption{Video-LLaVA performance in Habitat testing scenes.}
\label{tab:habitat-videollava-appendix}
\resizebox{0.5\textwidth}{!}{
\begin{tabular}{@{}lcccc@{}} 
\toprule
\textbf{Scene} & \textbf{SR(\%)$\uparrow$} & \textbf{SPL(\%)$\uparrow$} & \textbf{TL$\downarrow$} & \textbf{NE$\downarrow$}\\ \midrule
2azQ1b91cZZ & 20.00 & 10.21 & 381.51 & 3.77 \\
8194nk5LbLH & 12.50 & 7.28 & 387.78 & 5.17 \\
EU6Fwq7SyZv & 14.00 & 7.92 & 385.03 & 4.04 \\
QUCTc6BB5sX & 19.00 & 12.24 & 378.96 & 5.26 \\
SN83YJsR3w2 & 20.00 & 12.34 & 379.00 & 3.72 \\
TbHJrupSAjP & 7.14 & 3.01 & 395.36 & 5.05 \\
UwV83HsGsw3 & 8.57 & 4.70 & 393.87 & 5.29 \\
Vt2qJdWjCF2 & 11.32 & 6.66 & 388.67 & 6.24 \\
VzqfbhrpDEA & 16.20 & 10.37 & 384.72 & 5.42 \\
X7HyMhZNoso & 11.27 & 4.48 & 393.52 & 4.49 \\
Z6MFQCViBuw & 24.60 & 15.43 & 371.56 & 4.64 \\
gTV8FGcVJC9 & 25.00 & 15.86 & 371.61 & 4.50 \\
gYvKGZ5eRqb & 10.00 & 7.19 & 389.55 & 5.23 \\
oLBMNvg9in8 & 14.00 & 9.90 & 380.34 & 4.40 \\
pLe4wQe7qrG & 13.33 & 12.94 & 377.50 & 3.17 \\
pa4otMbVnkk & 10.00 & 5.57 & 387.50 & 6.02 \\
rqfALeAoiTq & 12.50 & 8.34 & 386.06 & 3.91 \\
x8F5xyUWy9e & 15.91 & 11.81 & 380.44 & 3.43 \\
yqstnuAEVhm & 10.54 & 6.06 & 390.36 & 5.19 \\
zsNo4HB9uLZ & 15.01 & 8.03 & 387.20 & 5.30 \\
\midrule
Average & 14.55 & 9.02 & 384.53 & 4.71 \\
\bottomrule
\end{tabular}
}
\end{table}

\begin{table}[htbp]\centering 
\caption{AVDC performance in Habitat testing scenes.}
\label{tab:habitat-avdc-appendix}
\resizebox{0.5\textwidth}{!}{
\begin{tabular}{@{}lcccc@{}} 
\toprule
\textbf{Scene} & \textbf{SR(\%)$\uparrow$} & \textbf{SPL(\%)$\uparrow$} & \textbf{TL$\downarrow$} & \textbf{NE$\downarrow$}\\ \midrule

2azQ1b91cZZ & 24.12 & 7.93 & 429.30 & 3.44 \\
8194nk5LbLH & 29.33 & 14.02 & 375.06 & 5.05 \\
EU6Fwq7SyZv & 26.95 & 12.35 & 442.17 & 5.45 \\
QUCTc6BB5sX & 22.14 & 10.48 & 450.81 & 3.54 \\
SN83YJsR3w2 & 30.77 & 15.06 & 395.85 & 4.21 \\
TbHJrupSAjP & 31.49 & 10.74 & 420.21 & 4.96 \\
UwV83HsGsw3 & 23.57 & 12.84 & 393.14 & 3.33 \\
Vt2qJdWjCF2 & 28.05 & 11.54 & 426.06 & 4.42 \\
VzqfbhrpDEA & 25.22 & 14.19 & 410.55 & 3.45 \\
X7HyMhZNoso & 29.58 & 12.82 & 414.27 & 4.20 \\
Z6MFQCViBuw & 26.78 & 13.78 & 430.31 & 4.63 \\
gTV8FGcVJC9 & 28.29 & 10.00 & 411.56 & 4.27 \\
gYvKGZ5eRqb & 26.52 & 10.18 & 421.96 & 4.92 \\
oLBMNvg9in8 & 25.44 & 13.00 & 417.07 & 3.91 \\
pLe4wQe7qrG & 27.88 & 16.95 & 429.00 & 4.40 \\
pa4otMbVnkk & 29.31 & 10.21 & 434.91 & 5.12 \\
rqfALeAoiTq & 28.45 & 12.56 & 403.33 & 4.42 \\
x8F5xyUWy9e & 24.68 & 9.76 & 432.66 & 3.68 \\
yqstnuAEVhm & 30.19 & 13.00 & 399.91 & 4.39 \\
zsNo4HB9uLZ & 26.94 & 13.32 & 424.30 & 4.14 \\
\midrule
Average & 27.29 & 12.24 & 418.12 & 4.30 \\
\bottomrule
\end{tabular}
}
\end{table}

\begin{table}[htbp]\centering 
\caption{VGM performance in Habitat testing scenes.}
\label{tab:habitat-vgm-appendix}
\resizebox{0.5\textwidth}{!}{
\begin{tabular}{@{}lcccc@{}} 
\toprule
\textbf{Scene} & \textbf{SR(\%)$\uparrow$} & \textbf{SPL(\%)$\uparrow$} & \textbf{TL$\downarrow$} & \textbf{NE$\downarrow$}\\ \midrule

2azQ1b91cZZ & 35.88 & 16.12 & 373.11 & 4.48 \\
8194nk5LbLH & 31.27 & 19.02 & 423.51 & 3.85 \\
EU6Fwq7SyZv & 34.10 & 18.74 & 368.62 & 4.06 \\
QUCTc6BB5sX & 36.27 & 16.88 & 367.95 & 4.29 \\
SN83YJsR3w2 & 32.92 & 17.09 & 426.29 & 4.73 \\
TbHJrupSAjP & 35.08 & 18.43 & 395.03 & 4.40 \\
UwV83HsGsw3 & 33.45 & 16.21 & 424.39 & 3.82 \\
Vt2qJdWjCF2 & 36.12 & 15.91 & 403.36 & 4.64 \\
VzqfbhrpDEA & 33.57 & 19.50 & 397.14 & 3.67 \\
X7HyMhZNoso & 34.88 & 18.62 & 398.75 & 4.05 \\
Z6MFQCViBuw & 32.67 & 15.53 & 398.10 & 4.37 \\
gTV8FGcVJC9 & 35.13 & 17.68 & 390.74 & 4.36 \\
gYvKGZ5eRqb & 34.77 & 16.44 & 393.34 & 4.36 \\
oLBMNvg9in8 & 31.22 & 17.13 & 391.16 & 3.38 \\
pLe4wQe7qrG & 36.45 & 18.04 & 357.49 & 4.19 \\
pa4otMbVnkk & 34.02 & 16.66 & 414.59 & 3.99 \\
rqfALeAoiTq & 35.23 & 18.21 & 369.25 & 4.20 \\
x8F5xyUWy9e & 32.45 & 17.90 & 380.19 & 4.32 \\
yqstnuAEVhm & 34.99 & 15.80 & 412.49 & 4.17 \\
zsNo4HB9uLZ & 36.23 & 16.09 & 392.92 & 4.39 \\
\midrule
Average & 34.34 & 17.30 & 393.92 & 4.19 \\
\bottomrule
\end{tabular}
}
\end{table}

\begin{table}[htbp]\centering 
\caption{ZSON performance in Habitat testing scenes.}
\label{tab:habitat-zson-appendix}
\resizebox{0.5\textwidth}{!}{
\begin{tabular}{@{}lcccc@{}} 
\toprule
\textbf{Scene} & \textbf{SR(\%)$\uparrow$} & \textbf{SPL(\%)$\uparrow$} & \textbf{TL$\downarrow$} & \textbf{NE$\downarrow$}\\ \midrule

2azQ1b91cZZ & 30.22 & 21.90 & 400.05 & 4.49 \\
8194nk5LbLH & 35.14 & 17.42 & 397.11 & 4.13 \\
EU6Fwq7SyZv & 31.88 & 18.15 & 360.23 & 4.71 \\
QUCTc6BB5sX & 34.62 & 21.22 & 379.75 & 4.42 \\
SN83YJsR3w2 & 30.87 & 19.88 & 415.70 & 4.21 \\
TbHJrupSAjP & 33.94 & 18.75 & 428.33 & 4.05 \\
UwV83HsGsw3 & 35.56 & 20.32 & 391.88 & 3.92 \\
Vt2qJdWjCF2 & 33.18 & 22.58 & 400.38 & 4.37 \\
VzqfbhrpDEA & 29.34 & 19.49 & 393.52 & 4.72 \\
X7HyMhZNoso & 34.90 & 18.22 & 376.97 & 4.29 \\
Z6MFQCViBuw & 32.12 & 21.74 & 380.55 & 4.91 \\
gTV8FGcVJC9 & 30.25 & 17.96 & 406.00 & 4.35 \\
gYvKGZ5eRqb & 35.44 & 22.08 & 412.43 & 3.81 \\
oLBMNvg9in8 & 31.53 & 19.14 & 366.82 & 4.58 \\
pLe4wQe7qrG & 30.87 & 20.69 & 393.75 & 4.14 \\
pa4otMbVnkk & 33.97 & 18.56 & 418.57 & 4.28 \\
rqfALeAoiTq & 34.12 & 19.24 & 378.22 & 4.33 \\
x8F5xyUWy9e & 32.78 & 18.97 & 416.39 & 3.91 \\
yqstnuAEVhm & 31.41 & 21.66 & 419.74 & 4.27 \\
zsNo4HB9uLZ & 34.55 & 19.31 & 382.41 & 4.43 \\
\midrule
Average & 32.83 & 19.86 & 395.94 & 4.32 \\
\bottomrule
\end{tabular}
}
\end{table}

\begin{table}[htbp]\centering \caption{NOLO-C performance in Habitat testing scenes.} \label{tab:habitat-nolo-c-appendix} \resizebox{0.5\textwidth}{!}{ \begin{tabular}{@{}lcccc@{}} \toprule \textbf{Scene} & \textbf{SR(\%)$\uparrow$} & \textbf{SPL(\%)$\uparrow$} & \textbf{TL$\downarrow$} & \textbf{NE$\downarrow$}\\ \midrule 2azQ1b91cZZ & 37.18$\pm$ \footnotesize{4.28} & 22.84$\pm$ \footnotesize{3.63} & 359.71$\pm$ \footnotesize{17.26} & 4.38$\pm$ \footnotesize{0.69} \\ 8194nk5LbLH & 36.97$\pm$ \footnotesize{1.55} & 19.47$\pm$ \footnotesize{0.99} & 366.57$\pm$ \footnotesize{4.66} & 2.97$\pm$ \footnotesize{0.34} \\ EU6Fwq7SyZv & 34.00$\pm$ \footnotesize{2.16} & 18.48$\pm$ \footnotesize{0.86} & 378.66$\pm$ \footnotesize{2.54} & 3.18$\pm$ \footnotesize{0.31} \\ QUCTc6BB5sX & 40.33$\pm$ \footnotesize{1.25} & 23.52$\pm$ \footnotesize{0.77} & 355.46$\pm$ \footnotesize{1.65} & 4.83$\pm$ \footnotesize{0.32} \\ SN83YJsR3w2 & 15.00$\pm$ \footnotesize{7.07} & 6.92$\pm$ \footnotesize{2.91} & 455.47$\pm$ \footnotesize{15.62} & 4.56$\pm$ \footnotesize{1.16} \\ TbHJrupSAjP & 20.95$\pm$ \footnotesize{5.99} & 6.05$\pm$ \footnotesize{1.97} & 437.43$\pm$ \footnotesize{14.11} & 4.53$\pm$ \footnotesize{0.51} \\ UwV83HsGsw3 & 26.67$\pm$ \footnotesize{1.35} & 17.00$\pm$ \footnotesize{1.76} & 401.74$\pm$ \footnotesize{3.26} & 4.77$\pm$ \footnotesize{0.41} \\
\bottomrule
\end{tabular} }
\end{table}

\begin{table}[htbp]\centering 
\caption{NOLO-T performance in Habitat testing scenes.}
\label{tab:habitat-nolo-t-appendix}
\resizebox{0.5\textwidth}{!}{
\begin{tabular}{@{}lcccc@{}} 
\toprule
\textbf{Scene} & \textbf{SR(\%)$\uparrow$} & \textbf{SPL(\%)$\uparrow$} & \textbf{TL$\downarrow$} & \textbf{NE$\downarrow$}\\ \midrule

2azQ1b91cZZ & 41.28$\pm$ \footnotesize{2.97} & 25.50$\pm$ \footnotesize{2.75} & 344.03$\pm$ \footnotesize{15.14} & 4.45$\pm$ \footnotesize{0.08} \\
8194nk5LbLH & 39.70$\pm$ \footnotesize{2.61} & 19.67$\pm$ \footnotesize{0.74} & 364.86$\pm$ \footnotesize{3.70} & 3.04$\pm$ \footnotesize{0.27} \\
EU6Fwq7SyZv & 41.00$\pm$ \footnotesize{3.27} & 21.33$\pm$ \footnotesize{2.90} & 357.03$\pm$ \footnotesize{11.76} & 3.37$\pm$ \footnotesize{0.17} \\
QUCTc6BB5sX & 41.50$\pm$ \footnotesize{4.50} & 21.52$\pm$ \footnotesize{2.39} & 357.77$\pm$ \footnotesize{11.40} & 5.51$\pm$ \footnotesize{0.73} \\
SN83YJsR3w2 & 21.67$\pm$ \footnotesize{11.79} & 7.83$\pm$ \footnotesize{4.20} & 432.92$\pm$ \footnotesize{32.98} & 5.79$\pm$ \footnotesize{1.89} \\
TbHJrupSAjP & 29.52$\pm$ \footnotesize{0.67} & 11.53$\pm$ \footnotesize{0.91} & 400.50$\pm$ \footnotesize{5.36} & 4.45$\pm$ \footnotesize{0.39} \\
UwV83HsGsw3 & 28.10$\pm$ \footnotesize{0.67} & 15.16$\pm$ \footnotesize{2.66} & 398.37$\pm$ \footnotesize{9.11} & 4.95$\pm$ \footnotesize{0.19} \\
\bottomrule
\end{tabular}
}
\end{table}

\begin{table}[htbp]\centering 
\caption{NOLO(M) performance in Habitat testing scenes.}
\label{tab:habitat-nolom-appendix}
\resizebox{0.5\textwidth}{!}{
\begin{tabular}{@{}lcccc@{}} 
\toprule
\textbf{Scene} & \textbf{SR(\%)$\uparrow$} & \textbf{SPL(\%)$\uparrow$} & \textbf{TL$\downarrow$} & \textbf{NE$\downarrow$}\\ \midrule

2azQ1b91cZZ & 35.77$\pm$ \footnotesize{0.38} & 21.70$\pm$ \footnotesize{0.98} & 367.77$\pm$ \footnotesize{0.44} & 5.26$\pm$ \footnotesize{0.06} \\
8194nk5LbLH & 38.18$\pm$ \footnotesize{0.91} & 19.20$\pm$ \footnotesize{0.74} & 365.92$\pm$ \footnotesize{3.05} & 3.05$\pm$ \footnotesize{0.18} \\
EU6Fwq7SyZv & 40.50$\pm$ \footnotesize{4.50} & 20.97$\pm$ \footnotesize{1.42} & 360.27$\pm$ \footnotesize{14.82} & 3.04$\pm$ \footnotesize{0.49} \\
QUCTc6BB5sX & 40.00 & 22.11$\pm$ \footnotesize{1.36} & 354.65$\pm$ \footnotesize{1.51} & 4.64$\pm$ \footnotesize{0.25} \\
SN83YJsR3w2 & 32.50$\pm$ \footnotesize{7.50} & 10.28$\pm$ \footnotesize{2.55} & 414.92$\pm$ \footnotesize{23.22} & 4.92$\pm$ \footnotesize{1.20} \\
TbHJrupSAjP & 24.29$\pm$ \footnotesize{1.43} & 7.48$\pm$ \footnotesize{0.50} & 427.30$\pm$ \footnotesize{6.51} & 4.63$\pm$ \footnotesize{0.07} \\
UwV83HsGsw3 & 30.71$\pm$ \footnotesize{2.14} & 17.89$\pm$ \footnotesize{3.05} & 390.06$\pm$ \footnotesize{15.81} & 4.03$\pm$ \footnotesize{0.21} \\
\bottomrule
\end{tabular}
}
\end{table}

%% file: tabs/cross.tex
\begin{table}[htbp]\centering 
\caption{NOLO performance in RoboTHOR after training in Habitat. ``T'' denotes FloorPlan\_Train and ``V'' denotes FloorPlan\_Val.}
\label{tab:rbothor-habitat-appendix}
\resizebox{0.5\textwidth}{!}{
\begin{tabular}{@{}lcccc@{}} 
\toprule
\textbf{Scene} & \textbf{SR(\%)$\uparrow$} & \textbf{SPL(\%)$\uparrow$} & \textbf{NE$\downarrow$} & \textbf{TL$\downarrow$}\\ \midrule

T10\_5 & 60.00$\pm$ \footnotesize{0.96} & 26.63$\pm$ \footnotesize{1.37} & 303.93$\pm$ \footnotesize{7.01} & 2.30$\pm$ \footnotesize{0.13} \\
T11\_5 & 64.89$\pm$ \footnotesize{3.50} & 31.85$\pm$ \footnotesize{3.64} & 286.70$\pm$ \footnotesize{20.00} & 2.09$\pm$ \footnotesize{0.22} \\
T12\_5 & 52.16$\pm$ \footnotesize{5.79} & 26.13$\pm$ \footnotesize{3.37} & 326.15$\pm$ \footnotesize{15.30} & 2.36$\pm$ \footnotesize{0.24} \\
T1\_5 & 50.00$\pm$ \footnotesize{6.24} & 29.40$\pm$ \footnotesize{1.07} & 320.91$\pm$ \footnotesize{10.47} & 2.46$\pm$ \footnotesize{0.04} \\
T2\_5 & 47.94$\pm$ \footnotesize{4.56} & 20.57$\pm$ \footnotesize{4.91} & 339.62$\pm$ \footnotesize{26.92} & 2.42$\pm$ \footnotesize{0.18} \\
T3\_5 & 39.22$\pm$ \footnotesize{3.09} & 18.48$\pm$ \footnotesize{0.60} & 377.95$\pm$ \footnotesize{6.26} & 3.30$\pm$ \footnotesize{0.22} \\
T4\_5 & 64.24$\pm$ \footnotesize{1.87} & 30.56$\pm$ \footnotesize{0.79} & 279.33$\pm$ \footnotesize{3.76} & 2.13$\pm$ \footnotesize{0.13} \\
\bottomrule
\end{tabular}
}
\end{table}

\begin{table}[htbp]\centering 
\caption{NOLO Performance in Habitat testing scenes after training in RoboTHOR.}
\label{tab:habitat-rbothor-appendix}
\resizebox{0.5\textwidth}{!}{
\begin{tabular}{@{}lcccc@{}} 
\toprule
\textbf{Scene} & \textbf{SR(\%)$\uparrow$} & \textbf{SPL(\%)$\uparrow$} & \textbf{NE$\downarrow$} & \textbf{TL$\downarrow$}\\ \midrule

2azQ1b91cZZ & 38.21$\pm$ \footnotesize{0.96} & 23.38$\pm$ \footnotesize{0.54} & 355.25$\pm$ \footnotesize{4.81} & 6.45$\pm$ \footnotesize{0.16} \\
8194nk5LbLH & 41.82$\pm$ \footnotesize{2.23} & 21.25$\pm$ \footnotesize{0.96} & 351.39$\pm$ \footnotesize{6.97} & 4.85$\pm$ \footnotesize{0.14} \\
EU6Fwq7SyZv & 45.67$\pm$ \footnotesize{2.87} & 22.27$\pm$ \footnotesize{1.75} & 345.99$\pm$ \footnotesize{10.11} & 4.72$\pm$ \footnotesize{0.37} \\
QUCTc6BB5sX & 44.00$\pm$ \footnotesize{2.45} & 23.97$\pm$ \footnotesize{1.79} & 346.11$\pm$ \footnotesize{11.27} & 6.91$\pm$ \footnotesize{0.19} \\
SN83YJsR3w2 & 18.33$\pm$ \footnotesize{4.71} & 7.28$\pm$ \footnotesize{0.33} & 449.10$\pm$ \footnotesize{9.22} & 9.20$\pm$ \footnotesize{0.77} \\
TbHJrupSAjP & 28.10$\pm$ \footnotesize{2.94} & 8.99$\pm$ \footnotesize{0.80} & 422.23$\pm$ \footnotesize{3.90} & 6.39$\pm$ \footnotesize{0.26} \\
UwV83HsGsw3 & 27.62$\pm$ \footnotesize{4.10} & 14.83$\pm$ \footnotesize{3.12} & 408.73$\pm$ \footnotesize{7.70} & 7.07$\pm$ \footnotesize{0.60} \\
\bottomrule
\end{tabular}
}
\end{table}